%% file: main.tex
\RequirePackage[svgnames]{xcolor}

\documentclass[11pt,letterpaper]{mystyle}

\usepackage[all]{hypcap}
\usepackage[svgnames]{xcolor}
\usepackage[comma,authoryear,compress]{natbib}
\usepackage{hyperref}[citecolor=lightblue]

\hypersetup{
    colorlinks = true,
    citecolor = {YaleBlue},
}

\usepackage{algorithm}
\usepackage{algorithmicx}
\usepackage{algpseudocode}
\usepackage{microtype}
\usepackage{graphicx}
\expandafter\def\csname ver@subfig.sty\endcsname{}
\usepackage{booktabs} %
\usepackage{float}
\usepackage{bigstrut}

\usepackage{amsmath}
\usepackage{amssymb}
\usepackage{mathtools}
\usepackage{amsthm}
\usepackage{mathrsfs}
\usepackage{nicefrac}
\usepackage{dsfont}
\usepackage{enumitem}
\usepackage{subcaption}
\usepackage{graphicx,subfig}
\usepackage{cleveref}
\usepackage{bxcoloremoji}

\usepackage{float}

\usepackage[utf8]{inputenc} %
\usepackage[T1]{fontenc}    %
\usepackage{hyperref}       %
\usepackage{url}            %
\usepackage{booktabs}       %
\usepackage{amsfonts}       %
\usepackage{nicefrac}       %
\usepackage{microtype}      %
\usepackage{graphicx}
\usepackage{subcaption} 
\usepackage{amssymb}
\usepackage{fdsymbol}
\usepackage{wrapfig}
\usepackage{lipsum}
\usepackage{enumitem}
\usepackage{stackengine}
\usepackage[font=small,labelfont=bf]{caption}
\usepackage{color}
\usepackage{adjustbox}

\usepackage{rotating}
\usepackage{makecell}

\usepackage{hyperref}

\usepackage{orcidlink}
\usepackage{makecell}
\usepackage{booktabs}
\usepackage{multirow}
\usepackage{multicol}
\usepackage{graphicx}
\usepackage[table]{xcolor} 
\usepackage{pifont}
\usepackage{textcomp}
\usepackage{wrapfig}
\usepackage{listings}
\usepackage{minted}
\usepackage{xcolor}
\usepackage{ragged2e}

\newcommand{\et}[2]{${#1}^{\pm{#2}}$}
\newcommand{\etb}[2]{$\mathbf{{#1}}^{\pm{#2}}$}
\newcommand{\ets}[2]{$\underline{{#1}}^{\pm{#2}}$}

\newcommand{\bluerow}{\rowcolor{cyan!6}}
\colorlet{baselinecolor}{cyan!4}
\newcolumntype{x}[1]{>{\centering\arraybackslash}p{#1pt}}
\newcolumntype{y}[1]{>{\raggedright\arraybackslash}p{#1pt}}
\newcolumntype{z}[1]{>{\raggedleft\array backslash}p{#1pt}}

\input{macro}

\newtcolorbox{AIbox}[2][]{aibox,title=#2,#1}
\definecolor{lightblue}{rgb}{0.22,0.45,0.70}%
\definecolor{Gray}{gray}{0.95}
\definecolor{Cornsilk}{rgb}{1.0, 0.97, 0.86}

\usepackage{amsmath}

\usepackage[all]{hypcap}

\title{Language-Guided Transformer Tokenizer for Human Motion Generation}

\runningtitle{Language-Guided Transformer Tokenizer for Human Motion Generation}

\author[1]{Sheng Yan}
\author[2]{Yong Wang}
\author[3]{Xin Du}
\author[4]{Junsong Yuan\orcidlink{0000-0002-7901-8793}}
\author[5]{Mengyuan Liu\orcidlink{0000-0002-6332-8316}\textsuperscript{\textdagger}}

\affil[1]{Transsion Ltd.}
\affil[2]{Chongqing University of Technology}
\affil[3]{Afari Intelligence Drive}
\affil[4]{State University of New York at Buffalo}
\affil[5]{State Key Laboratory of General Artificial Intelligence, Peking University, Shenzhen Graduate School}

\correspondingauthor{$\dagger$ Mengyuan Liu \href{mailto:liumengyuan@pku.edu.cn}{liumengyuan@pku.edu.cn}}

\begin{document}

\input{sec/0_abstract}

\maketitle  
\input{sec/1_intro}
\input{sec/2_related_work}

\input{sec/3_method}
\input{sec/4_experiment}

\input{sec/5_conclusion}
\input{sec/6_acknowledge}

\clearpage
\bibliography{main}

\appendix
\input{sec/X_suppl}
\appendix
\end{document}

%% file: macro.tex
\newcommand{\x}{\mathbf{x}}

\newcommand{\etal}{{et~al.}\ }
\newcommand{\eg}{\emph{e.g.},\ }
\newcommand{\ie}{\emph{i.e.},\ }

\def\etc{\emph{etc.}\ }

\definecolor{blanchedalmond}{rgb}{1.0, 0.92, 0.8}
\definecolor{carmine}{rgb}{0.59, 0.0, 0.09}
\definecolor{lightblue}{rgb}{0.22,0.45,0.70}%

\renewcommand{\mathbf}{\boldsymbol}

\makeatletter
\def\Ddots{\mathinner{\mkern1mu\raise\p@
\vbox{\kern7\p@\hbox{.}}\mkern2mu
\raise4\p@\hbox{.}\mkern2mu\raise7\p@\hbox{.}\mkern1mu}}
\makeatother

\definecolor{amaranth}{rgb}{0.9, 0.17, 0.31}
\definecolor{antiquebrass}{rgb}{0.8, 0.58, 0.46}
\definecolor{antiquefuchsia}{rgb}{0.57, 0.36, 0.51}
\definecolor{chromeyellow}{rgb}{0.31, 0.47, 0.26}

\newcommand{\github}{\raisebox{-1.5pt}{\includegraphics[height=1.05em]{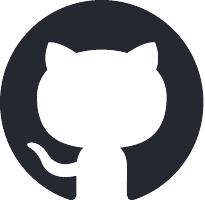}}}

%% file: sec/0_abstract.tex
\begin{abstract}
In this paper, we focus on motion discrete tokenization, which converts raw motion into compact discrete tokens—a process proven crucial for efficient motion generation. In this paradigm, increasing the number of tokens is a common approach to improving motion reconstruction quality, but more tokens make it more difficult for generative models to learn.
To maintain high reconstruction quality while reducing generation complexity, we introduce Language-Guided Tokenization (LG-Tok) for efficient motion tokenization. LG-Tok aligns natural language with motion at the tokenization stage, yielding compact, high-level semantic representations. This approach not only strengthens both tokenization and detokenization but also simplifies the learning of generative models. Furthermore, existing tokenizers predominantly adopt convolutional architectures, whose local receptive fields struggle to support global language guidance. To this end, we propose a Transformer-based Tokenizer that leverages attention mechanisms to enable effective alignment.
Additionally, we design a language-drop scheme, in which language conditions are randomly removed during training. This scheme prevents shortcut learning over text and enables the detokenizer to support language-free guidance. On three generation benchmarks, LG-Tok outperforms state-of-the-art methods (e.g., achieving an FID score of 0.057 vs. MARDM's 0.114 on HumanML3D). LG-Tok-mini uses only half the tokens while maintaining competitive performance, validating the efficiency of our semantic representations.

\vspace{2mm}


\vspace{5mm}

\coloremojicode{1F4C5} \textbf{Date}: June 30, 2026

\coloremojicode{1F3E0} \textbf{Projects}: \href{https://eanson023.github.io/LG-Tok/}{https://eanson023.github.io/LG-Tok/}

\github{} \textbf{Code Repository}: \href{https://github.com/eanson023/LG-Tok}{https://github.com/eanson023/LG-Tok}





\end{abstract}

%% file: sec/1_intro.tex
\section{Introduction}
\label{sec:intro}

Text-driven human motion generation~\cite{zhu2023human, petrovich2021action,guo2020action2motion,ghosh2021synthesis,
tang2018dance, le2023music, ginosar2019learning, alexanderson2023listen,
zhu2023taming}
enables the synthesis of realistic human motions based on natural language
descriptions, with widespread applications in game animation, virtual reality, and
video motion editing, etc. In recent years, the two-stage paradigm, combining motion
tokenization with generative models~\cite{zhang2023generating, pinyoanuntapong2024mmm,
guo2024momask, yuan2024mogents, zhang2025motion, wang2025spatial}, has
demonstrated a pivotal role in efficiently synthesizing high-fidelity motions. These
methods tokenize continuous motion representations into discrete tokens, thereby
enabling the direct utilization of well-established generative transformer
training and sampling techniques~\cite{li2024autoregressive, tian2024visual,brown2020language,radford2019language,chang2022maskgit}
with minimal modifications.

\begin{figure}[htbp]
    \centering
    \includegraphics[width=0.6\textwidth, trim=0mm 0mm 3mm 0mm, clip]{
        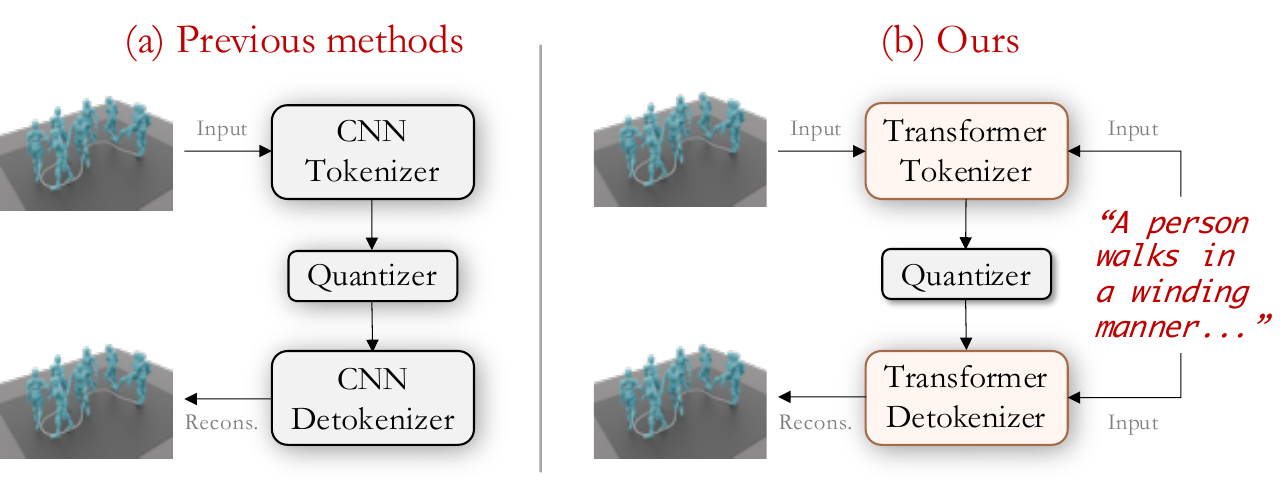
    }
    \vspace{-10pt}
    \caption{Comparison between previous CNN-based tokenizers and our Language-Guided
    Transformer Tokenizer (LG-Tok). Our method aligns language and motion during
    tokenization, leveraging the transformer's flexibility.}
    \vspace{-10pt}
    \label{fig:teaser}
\end{figure}

\begin{wraptable}[6]
    {r}{0.5\textwidth}
    \centering
    \vspace{0.2cm}
    \scalebox{1.0}{
    \begin{tabular}{c|c|c|c}
        \#Tokens         & 104   & 160   & 236   \\
        \hline
        rFID$\downarrow$ & 0.143 & 0.110 & 0.090 \\
        \hline
        gFID$\downarrow$ & 0.230 & 0.205 & 0.257 \\
    \end{tabular}}
    \vspace{-0.3cm}
    \caption{Performance against \#Tokens. rFID/gFID measure reconstruction/generation quality.}
    \label{tab:intro_tradeoff}
\end{wraptable}
As the core of this paradigm, the properties of motion tokenization significantly
influence the performance of generative transformers. Despite various motion
tokenization studies focusing on improving training objectives~\cite{guo2022tm2t,zhang2023generating}
or optimizing quantization~\cite{guo2024momask, lu2025scamo, liu2025mosa}, current
methods still suffer from a fundamental trade-off between reconstruction and generation
quality. As shown in Table~\ref{tab:intro_tradeoff}, increasing the number of tokens typically
improves motion reconstruction quality (rFID). However, the generation process exhibits
a different behavior—more tokens increase learning difficulty, leading to
suboptimal results (gFID). Based on this observation, we raise the question: \textit{Can
we maintain sufficient tokens for high-quality reconstruction while
simultaneously reducing the learning difficulty of token sequences for
generative models?}

Our key insight is that language descriptions naturally provide high-level semantic
abstractions—\eg, \textit{``a person walks forward''} encapsulates core motion
intent. By introducing language as auxiliary guidance during tokenization, tokens
are relieved from redundantly encoding high-level semantics and can instead focus
on capturing the fine-grained motion details that language alone cannot fully convey. To this end, we propose Language-Guided Tokenization (LG-Tok), which aligns
natural language with motion during tokenization, aiming to provide compact, high-level
semantic representations. This approach, typically reserved for the generation phase,
is extended to the tokenization stage in our work. It achieves three benefits:
\textit{1)} enables latent tokens to learn motion representations while simultaneously
absorbing linguistic semantic knowledge during the tokenizing process; \textit{2)}
allows language conditions to assist in effectively reconstructing the input motion
during the detokenizing process. \textit{3)} simplifies the learning of generative
models. For instance, on the Motion-X dataset~\cite{lin2023motion}, our compact
semantic representations reduce model perplexity from 146.5 to 103.1 and improve
FID from 0.257 to 0.088, indicating easier and more effective learning.

Notably, existing tokenizers~\cite{guo2022tm2t, zhang2023generating,
guo2024momask, liu2025mosa}
predominantly adopt convolutional tokenizer-detokenizer architectures, whose local receptive fields struggle to support global language guidance. To address this, we propose a Transformer-based Tokenizer that leverages flexible attention mechanisms to enable effective alignment between language and motion, while enhancing the global contextual awareness of token representations. Specifically, we predefine a learnable sequence of latent tokens that are concatenated with both motion and language representations. Following feature encoding by the transformer-based tokenizer, these latent tokens form semantically informed representations. After vector quantization, a transformer-based detokenizer reconstructs the input motion from the masked token sequences.

\begin{wrapfigure}[9]
    {r}{0.5\columnwidth}
    \centering
    \vspace{-14pt}
    \includegraphics[width=0.45\columnwidth, trim=2mm 0mm 6mm 0mm, clip]{
        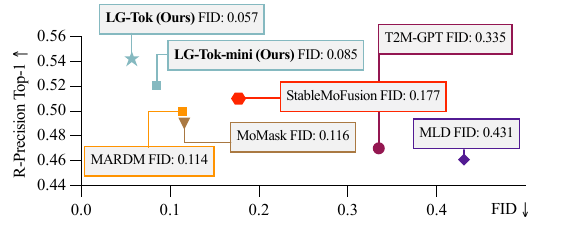
    }
    \vspace{-10pt}
    \caption{\small Generation quality on HumanML3D.}
    \label{fig:teaser2}
\end{wrapfigure}

Furthermore, we observe that naively incorporating language conditions risks shortcut learning~\cite{geirhos2020shortcut}, where the tokenizer-detokenizer tends to over-rely on well-trained text embeddings for encoding and reconstruction, undermining the tokens' ability to capture fine-grained motion details. To mitigate this, we propose a language-drop scheme that randomly removes language conditions during training, compelling the model to develop robust token representations independent of linguistic input. Beyond preventing shortcut learning, this scheme further enables language-free guidance decoding during generation.
Specifically, drawing on insights from Classifier-Free Guidance~\cite{ho2022classifier} for generative models operating in the logit or noise space, our approach extends this principle directly to the \textit{motion space}, providing a complementary guidance mechanism.
Fig.~\ref{fig:teaser2} shows our LG-Tok outperforming representative baselines. We evaluate LG-Tok on three generation benchmarks: HumanML3D~\cite{guo2022generating}, KIT-ML~\cite{plappert2016kit}, and Motion-X.
Across all datasets, LG-Tok consistently surpasses state-of-the-art methods, \eg, achieving a Top-1 R-Precision of \textbf{0.542} and FID of \textbf{0.057} on HumanML3D, outperforming MARDM~\cite{meng2025rethinking} (0.500 and 0.114).
Meanwhile, LG-Tok-mini, using only half the tokens while maintaining competitive performance (Top-1: 0.521, FID: 0.085 on HumanML3D), validates the efficiency of our semantic representations. In summary, our contributions are as follows:
\begin{itemize}
    \item We propose Language-Guided Tokenization (LG-Tok), which achieves alignment between natural language and motion during the motion tokenization phase.

    \item We introduce a Transformer-based Tokenizer that leverages attention mechanisms to enable effective language-motion alignment and enhance global contextual awareness of token representations.

    \item We propose a language-drop scheme that prevents shortcut learning and further enables language-free guidance decoding in the motion space during generation.
\end{itemize}

%% file: sec/2_related_work.tex
\section{Related Work}
\label{sec:related_work}

\noindent\textbf{Motion tokenization.} Motion tokenizers~\cite{zhang2023generating, guo2022tm2t, guo2024momask, pinyoanuntapong2024mmm, pinyoanuntapong2024bamm, petrovich2022temos, wu2024motionllm, komura2017recurrent, guo2020action2motion} play a crucial role in efficiently synthesizing high-fidelity motion by reducing computational costs while improving generation quality and efficiency. This has been achieved through autoencoders (AEs)~\cite{komura2017recurrent, tevet2022motionclip}. This general design compresses motion into low-dimensional representations, which are decoded back to the original space. Variational autoencoders (VAEs)~\cite{kingma2013auto, kingma2019introduction} extend this paradigm by introducing probability distributions, enabling stochastic sampling and generative capabilities. TEMOS~\cite{petrovich2022temos} is a representative work, aligning a text encoder’s output distribution with the VAE’s latent space to enable text-conditioned motion generation. Latent diffusion models~\cite{chen2023executing, andreou2025lead,azadi2023make} further exploit compressed spaces—avoiding the cost of diffusion in the raw motion space. More recently, MARDM~\cite{meng2025rethinking} performs masked autoregressive diffusion directly on deterministic AE latents, mitigating the sampling noise inherent to VAE representations.

Alternatively, Vector Quantized VAEs (VQ-VAE)~\cite{van2017neural} insert a quantization step between AE encoders (tokenizers) and decoders (detokenizers), replacing continuous latent distributions with discrete codes that partition the motion space into categorical units. TM2T~\cite{guo2022tm2t} first applied VQ-VAE to motion, introducing discrete motion tokens. This discrete tokenization facilitates the use of powerful sequence-based generative models~\cite{bahdanau2014neural, radford2019language, devlin2019bert, chang2022maskgit}. Subsequent variants~\cite{zhang2023generating,zhang2025motion,shi2025genm}, \eg RQ-VAE~\cite{guo2024momask,zhang2024kmm} and FSQ~\cite{wang2025spatial,lu2025scamo}, further refined the quantization and enabled the integration of advanced generative architectures. In this paper, we focus on this discrete (\ie,  VQ-based) tokenization pipeline. Based on our observation of the fundamental trade-off between reconstruction and generation in current methods, we propose leveraging language guidance during tokenization to provide compact semantic representations, thereby maintaining high reconstruction quality while reducing the learning difficulty for generative models.

\noindent\textbf{Text-driven motion generation.} 
Natural language provides rich semantic cues for specifying actions, velocities, and directions, making it a key conditioning modality for human motion synthesis~\cite{petrovich2021action,guo2020action2motion,ghosh2021synthesis, tang2018dance, le2023music, ginosar2019learning, alexanderson2023listen, zhu2023taming, zhai2023language}. Early GAN-based work, such as Text2Action~\cite{ahn2018text2action}, generated diverse motions from textual descriptions, while JL2P~\cite{ahuja2019language2pose} employed GRU-based encoders and decoders to map text to movement. Inspired by advances in text-to-image generation, recent approaches predominantly adopt diffusion or VQ-based paradigms. Diffusion models~\cite{zhang2023remodiffuse, zhang2022motiondiffuse,karunratanakul2023guided,zhou2024emdm, kim2023flame, wan2023diffusionphase, lou2023diversemotion} simulate a forward noising process and train networks to reverse it; MLD~\cite{chen2023executing} improves efficiency by operating in latent space, while PhysDiff~\cite{yuan2023physdiff} enforces physical constraints during generation. VQ-based methods~\cite{yuan2024mogents,zhong2023attt2m,li2024lamp,zhang2025motion,shi2025genm,zhang2024kmm,wang2025spatial,lu2025scamo,pinyoanuntapong2024bamm}, \eg, T2M-GPT~\cite{zhang2023generating}, quantize motion into discrete tokens via VQ-VAE and model them autoregressively using GPT-style~\cite{brown2020language,radford2019language} transformers. MoMask~\cite{guo2024momask} and MMM~\cite{pinyoanuntapong2024mmm} adopt MaskGIT-style~\cite{chang2022maskgit} masked training and non-autoregressive sampling, and MoSa~\cite{liu2025mosa} adapts scale-wise autoregressive modeling from the image domain~\cite{tian2024visual}. In this paper, we adopt MoSa as our generative model given its notable gains in quality and efficiency. To demonstrate the generalizability of our approach, we also apply LG-Tok to MoMask, showing consistent improvements across different generative paradigms.



\noindent\textbf{Image tokenization} has emerged as a fundamental technique bridging various vision tasks. This field has diverged into two main branches: understanding-oriented and generation-oriented. Understanding-oriented approaches~\cite{ma2025unitok,li2023blip,alayrac2022flamingo} leverage large language models (LLMs) to create semantic representations for tasks such as classification~\cite{dosovitskiy2020image} and segmentation~\cite{wang2021max}. Meanwhile, generation-oriented methods, such as VQ-GAN~\cite{esser2021taming,razavi2019generating}, emphasize learning latent spaces for detail-preserving compression. These methods typically employ 2D latent grids with fixed downsampling factors~\cite{zhu2023taming,lee2022autoregressive,tian2024visual,chang2022maskgit,li2024imagefolder,qiu2025robust}. TikTok~\cite{yu2024image} first introduces a transformer-based 1D tokenizer that generates global tokens as more compact representations of images. Subsequently, numerous variants have refined it~\cite{wu2025alitok,chen2025softvq,zhu2024scaling,yang2025latent}. \eg, TxtTok~\cite{zha2025language} extends this paradigm by incorporating image captions during tokenization to achieve higher compression rates. While we share the high-level concept of language guidance, our technical design fundamentally departs from TxtTok. Naive text conditioning risks representation collapse (\ie, shortcut learning) and restricts decoding flexibility. Instead, a key design of ours is a language-drop scheme—absent in TxtTok—that not only robustifies discrete representations but uniquely empowers the detokenizer to execute classifier-free guidance directly in the \textit{motion space} during generation. Furthermore, we propose a fully transformer-based architecture, while been explored in continuous VAE representations~\cite{chen2023executing,petrovich2022temos}, their application in VQ-based tokenizers remains limited—existing attempts~\cite{guo2025snapmogen,liu2025mosa} only stack shallow self-attention layers after CNN-based residual blocks, which may be insufficient for effective cross-modal alignment. Moreover, while M2DM~\cite{kong2023priority} also adopts a Transformer-based VQ-VAE, it shares with prior motion tokenizers a 1:1 frame-to-latent alignment trained on 64-frame short crops. In contrast, LG-Tok encodes 196-frame sequences via $N$ learnable queries fully decoupled from frame count ($N{=}25$ in LG-Tok-mini), enabling each token to capture sentence-level semantics rather than local per-frame features.

%% file: sec/3_method.tex
\section{Method}
\label{sec:method}

\begin{figure}[!t]
    \centering
    \vspace{-10pt}
    \includegraphics[width=1.0\columnwidth]{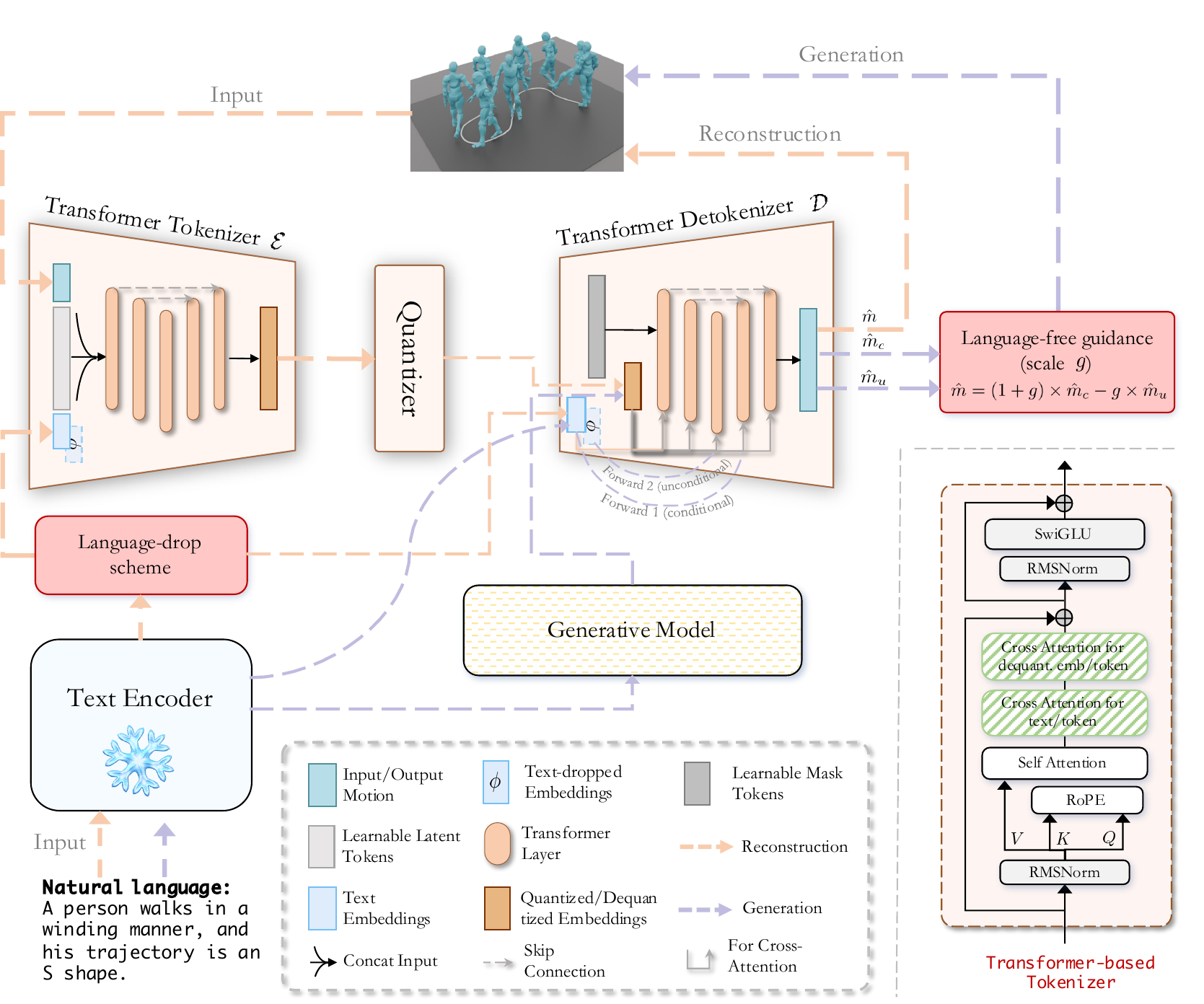}
    \vspace{-15pt}
    \caption{\textbf{Overview of LG-Tok.} Given a motion sequence and its natural language description, a frozen text encoder extracts text embeddings, which are concatenated with learnable latent tokens and linearly projected motion as inputs to the Transformer-based tokenizer $\mathcal{E}$ (self-attention only), producing discrete motion tokens via the quantizer. A language-drop scheme randomly discards text input with probability $p$, preventing shortcut learning. In the Transformer-based detokenizer $\mathcal{D}$, learnable mask tokens interact with text embeddings and dequantized embeddings via two dedicated cross-attention layers (shown on the right), respectively, to reconstruct the motion sequence.
    For Generation, motion tokens sampled by the generative model are dequantized and fed into $\mathcal{D}$ with language-free guidance—two forward passes (conditional and unconditional) are performed and extrapolated in motion space to synthesize diverse, high-fidelity human motion.}
    \label{fig:model}
\end{figure}

\subsection{Preliminary}
\label{subsec:preliminary}

\noindent
\textbf{Motion tokenizer.} Human motion tokenization converts continuous motion
representations into discrete tokens to facilitate the generative models. Traditional
approaches employ vanilla VQ-VAE~\cite{guo2022tm2t, zhang2023generating,
pinyoanuntapong2024mmm}, where a tokenizer (encoder) $\mathcal{E}(\cdot)$
compresses motion $m$ into latent features
$z = \mathcal{E}(m) \in \mathbb{R}^{T \times d}$, followed by a quantizer $\mathcal{Q}
(z)$ that maps each latent feature to its nearest codebook entry, producing
discrete tokens $x \in [V]^{T}$, where $V$ is the codebook size. Then, a detokenizer
(decoder) $\mathcal{D}(\cdot)$ reconstructs motion from the dequantized embeddings
$\hat{m}= \mathcal{D}(\hat{z})$. To address quantization errors, Residual VQ-VAE~\cite{guo2024momask,zhang2024kmm}
performs iterative residual quantization with $N$ quantizers, creating the same-scale
token sets $(x^{(1)}, \dots, x^{(N)})$ where each $x^{(n)}\in [V]^{T}$. Recent work,
MoSa~\cite{liu2025mosa}, introduces interpolations before each residual quantization
to construct tokens at different scales $S = (s_{1}, s_{2}, \dots, s_{N})$ through
downsampling: $x^{(n)}= \mathcal{Q}^{(n)}({\big\Downarrow}(z^{(n)}, s_{n}))$
where the symbol ${\big\Downarrow}(\cdot, s_{n})$ denotes downsampling latent
features to specific granularities $s_{n}$. This process produces compact multi-scale
tokens $x^{(n)}\in [V]^{s_n}$ rather than same-scale representations.


\noindent
\textbf{Generative model.} Leveraging multi-scale tokens, MoSa enables Scale-wise
Autoregressive (SAR) modeling that reformulates traditional token-by-token
prediction~\cite{zhang2023generating} into scale-by-scale generation. The SAR likelihood
is defined as:
\begin{small}
    \begin{align}
        p(x^{(1)}, \dots, x^{(N)}\mid c) = \prod_{n=1}^{N}p(x^{(n)}\mid x^{(1)}, \dots, x^{(n-1)}, c)
    \end{align}
\end{small}

\noindent
where $x^{(n)}= (x^{(n)}_{1}, \dots, x^{(n)}_{s_n})$ represents all tokens at scale
$s_{n}$, and will be predicted simultaneously at step $n$. This approach generates
multiple tokens in parallel at each autoregressive step, conditioned on previous
scales and condition $c$. Given this framework's notable gains in quality and efficiency,
we adopt it as our generative model. In the following sections, we use LG-Tok to
refer to this complete generation framework by default. We also apply LG-Tok to MoMask to demonstrate
its generalizability. The complete tokenization-generation-detokenization
pipeline is detailed in the Appendix~\ref{sec:sub_tok_gen_detok_pipeline}.

\subsection{Transformer-based Tokenizer}
\label{subsec:transformer}

To facilitate understanding of our approach, we introduce our Transformer-based Tokenizer first.

Existing discrete tokenizers have achieved substantial results but still exhibit a limitation in their standard workflow: The 196-frame motion is compressed into 49 latent tokens via 1D convolutional encoding with 4× downsampling, whose local receptive fields struggle to support global language guidance and to enhance the global contextual awareness of token representations. To this end, we propose a Transformer-based Tokenizer. As depicted in Fig.~\ref{fig:model}, both our tokenizer and detokenizer are attention-based~\cite{vaswani2017attention}. We predefine a learnable sequence of latent tokens and use these tokens for reconstruction and subsequent generation. With the self-attention mechanism, token representations can enhance global contextual awareness. After feature encoding, these latent tokens constitute a representation of motion. Specifically, we concatenate a set of learnable tokens $z_{l}\in \mathbb{R}^{T\times d}$ of length $T$ with linearly transformed motion $m \in \mathbb{R}^{F\times d}$ as the input to the tokenizer, and retain only the output corresponding to the learnable latent tokens as the tokenizer output:
\begin{equation}
    z = \mathcal{E}([z_{l};m]) \label{eq:encoder}
\end{equation}
subsequently, the latent tokens undergo vector quantization to obtain multi-scale discrete tokens (as mentioned in Sec.~\ref{subsec:preliminary}) to support subsequent SAR modeling, and yield dequantized embeddings $\hat{z}\in \mathbb{R}^{T\times d}$. For the detokenizer, we reconstruct motion from a sequence of learnable mask tokens $\hat{m}_{l}\in \mathbb{R}^{F\times d}$~\cite{he2022masked,devlin2019bert}:
\begin{equation}
    \hat{m}= \mathcal{D}(\hat{m}_{l}, \hat{z})
\end{equation}
unlike the tokenizer, the mask tokens interact with the dequantized embeddings through cross-attention. This design is inspired by object queries~\cite{carion2020end} in object detection, and we provide detailed ablations in the experimental section. At the detokenizer end, we use a linear layer to regress from mask tokens to motion space. Notably, we do not apply patchify processing~\cite{dosovitskiy2020image} to motion throughout process, since 1D motion (196 frames) incurs a
significantly lower computational cost than 2D images (256×256), and this operation also also avoids information loss.

Our architecture is closely integrated with LLaMA. We incorporate RMSNorm~\cite{zhang2019root}, SwiGLU activation~\cite{shazeer2020glu}, and advanced rotary position embedding (RoPE)~\cite{su2024roformer}. The RoPE \textit{base} is set to 100 to accommodate short sequence tasks. We further enhance the transformer networks of both tokenizer $\mathcal{E}$ and detokenizer $\mathcal{D}$ with UNet-like long skip connections for higher-fidelity motion reconstruction.

\subsection{Language-Guided Tokenization}
\label{subsec:language}

Introducing language as auxiliary guidance during tokenization, tokens are relieved from redundantly encoding high-level semantics and can instead focus on capturing the fine-grained motion details that language alone cannot fully convey. Building on the flexibility of our transformer-based tokenizer, we introduce Language-Guided Tokenization, which aligns natural language with motion during tokenization. This approach, typically reserved for the generation phase, is extended to the tokenization stage in our work. Given textual descriptions of motion, we use a frozen LLaMA~\cite{grattafiori2024llama} as the text encoder to extract text embeddings. These embeddings are injected into both the tokenizer and detokenizer, providing semantic guidance throughout the tokenization process. As illustrated in Fig.~\ref{fig:model}, the tokenizer
input is further concatenated with linearly projected text embeddings
$t \in \mathbb{R}^{W \times d}$. The Eq.~\ref{eq:encoder} is updated to
$z = \mathcal{E}([t;z_{l};m])$, yielding compact, high-level semantic
representations. For the detokenizer, the mask tokens additionally interact with text embeddings through another cross-attention: $\hat{m}= \mathcal{D}(\hat{m}_{l}, \hat{z}, t)$. We train LG-Tok using smooth L1 loss for motion reconstruction, without involving text reconstruction. In the generation phase, the provided text descriptions are used for both generation and detokenization. The text embeddings and the latent tokens sampled by the generative model are fed into the detokenizer to produce the final motion.

Despite its simplicity, we emphasize that this approach achieves three benefits:
\textit{1)} enables latent tokens to learn motion representations while simultaneously absorbing linguistic semantic knowledge during the tokenizing process; \textit{2)} allows language conditions to assist masked token sequences in effectively reconstructing the input motion during the detokenizing process. \textit{3)} simplifies the learning of generative models by more compact semantic representations. We demonstrated these views in our experiments.

\vspace{-3pt}
\subsection{Language-Drop Scheme}
\label{subsec:language-drop}
\vspace{-3pt}

While language guidance benefits tokenization, naively incorporating it risks shortcut learning: the tokenizer-detokenizer may over-rely on well-trained text embeddings for encoding and reconstruction, undermining the tokens' ability to capture fine-grained motion details.
To mitigate this, we propose a language-drop scheme that randomly removes language conditions during training.
Specifically, we drop the text input ($t=\emptyset$) with a probability of $p$, compelling the model to develop robust token representations independent of linguistic input.
This ensures that the learned tokens remain informative and self-sufficient, rather than collapsing into mere pointers to semantic text features.

Beyond preventing shortcut learning, this scheme enables language-free guidance decoding during the token-to-motion generation phase.
Since the detokenizer is trained to operate both with and without text, we can apply Classifier-Free Guidance directly at the final detokenizing stage. The final generated motion $\hat{m}$ is obtained by extrapolating from the unconditional motion $\hat{m}_{u}$ to the conditional motion $\hat{m}_{c}$ with guidance scale $g$:
\begin{equation}
    \hat{m}= (1+g)\times\hat{m}_{c}-g\times\hat{m}_{u}
\end{equation}
This provides a complementary guidance mechanism to existing methods that operate in the logits space. Our approach further applies guidance at the motion space, offering a lightweight yet effective boost to generation quality.

%% file: sec/4_experiment.tex
\section{Experiment}
\label{sec:experiment}

In this section, we present the results of our experiments. We introduce our experimental setup in Sec. \ref{sec:experiment_setup}. Subsequently, we compare our results with competing methods in Sec. \ref{sec:comparision} and \ref{sec:vq_tok}, followed by an in-depth analysis of the language-drop scheme in Sec. \ref{sec:analysis_of_language_drop} and related ablation experiments in Sec. \ref{sec:ablation}. Finally, we provide reconstruction–generation trade-off analysis in Sec.~\ref{sec:tradeoff} and tokenizer analysis in Sec. \ref{sec:tokenizer_analysis}. Additional discussions on the extended applications, inference time, limitations, and more are provided in the Appendix.

\subsection{Experimental Setup}
\label{sec:experiment_setup}

We conduct experiments on three text-to-motion benchmarks: 
HumanML3D~\cite{guo2022generating}, KIT-ML~\cite{plappert2016kit} and the larger-scale Motion-X dataset~\cite{lin2023motion}. We follow the most evaluation protocol proposed in \cite{guo2022generating, meng2025rethinking}.  

\noindent\textbf{Datasets.} HumanML3D dataset includes 14,616 high-quality motions paired with 44,970 text descriptions, where three different captions describe each motion. KIT-ML comprises 3,911 motions with 6,278 text annotations. Motion-X is the larger motion-text dataset, featuring greater diversity. Following the protocol of the first dataset, we filter out motion-text pairs exceeding 200 frames, resulting in 37,751 motion sequences and 61,637 text captions. The datasets are split into training, validation, and test sets with ratios of 80\%, 5\%, and 15\%, respectively.  

For standardization, all datasets adopt the \texttt{meng} (64-67 dim) representation~\cite{meng2025rethinking}, as their research uncovered the redundancy of the original features. That is, Motion-X's whole-body representation is also converted to \texttt{meng}. Since most textual descriptions primarily focus on body actions, we choose to ignore finger and facial information to avoid unnecessary modal discrepancies. We provide supplementary training details for the Motion-X feature extractor in Appendix~\ref{sec:feature_extractor_training}.

\noindent\textbf{Evaluation metrics.} We adopt the following evaluation metrics:  
(1) the \textit{Frechet Inception Distance (FID)}, which assesses the overall action quality by measuring the distributional difference between the high-level features of generated and real actions; 
(2) \textit{R-Precision} and \textit{Multimodal distance}, which are used to measure the semantic consistency between the input text and the generated actions; 
(3) \textit{Multimodality}, which is used to evaluate the diversity of actions generated from the exact text.
(4) \textit{CLIP-score}, which measures the compatibility of motion-text pairs by calculating the cosine similarity.

\input{table/table1}

\noindent\textbf{Implementation details.} We train LG-Tok using the AdamW optimizer with a batch size of 128 for 200 epochs. The learning rate is reduced from $2\times 10^{-4}$ to $2\times 10^{-5}$ at the 180th epoch. We set the gradient clipping factor to 0.01 to prevent gradient explosion. No velocity loss is employed during optimization, as the \texttt{meng} representation is sufficiently compact. We set the probability $p=0.1$ of our language-drop scheme. For the Transformer, we stack 9 layers for both the tokenizer and detokenizer, each with 4 heads, 256 latent dimensions, and a SwiGLU dimension of 1024. To accelerate training and reduce memory usage, we enable mixed-precision training and utilize PyTorch 2.2.0 to support flash attention. For text guidance, \textit{LLaMA-3.2-1B} serves as our text encoder, and the maximum text length is set to 77, with more extended sequences truncated. The guidance scale $g$ is set to 2.0, 1.0 and 2.0 on the HumanML3D, Motion-X and KIT-ML, respectively. During generation, the generative model's configuration remains unchanged, except that \texttt{PAD} masks are no longer required. Note that the generative model and the evaluation retrieval metrics independently use \textit{CLIP-ViT-B/32} for text embedding, consistent with all baselines; \textit{LLaMA-3.2-1B} features are exclusive to LG-Tok's tokenizer and detokenizer. LG-Tok training can be completed on a single RTX 4090 GPU.

\noindent\textbf{Model variants.} To validate the high-level semantic representations of LG-Tok, we study three variants with different token scales: LG-Tok-mini (25 tokens), LG-Tok-mid (36 tokens), and LG-Tok (49 tokens). The interpolation scales mentioned in Sec.~\ref{subsec:preliminary} are set to $S=(1,2,\dots,25)$, $S=(2,4,\dots,36)$, and $S=(3,6,\dots,49)$ respectively. All variants use $N=10$ scales (quantizers), resulting in 104, 160, and 236 total tokens, respectively.

\subsection{Text-driven Motion Generation Comparison}
\label{sec:comparision}

\noindent\textbf{Quantitative comparisons.} We evaluate our approach against state-of-the-art methods. Results on HumanML3D are primarily sourced from~\cite{meng2025rethinking}, while results on Motion-X are our reproduction based on the \texttt{meng} representation. We provide reproduction details in the Appendix~\ref{sec:baseline_reprod}. As shown in Table~\ref{tab:table1}, our LG-Tok variants (using MoSa as the generative backbone) outperform existing methods across multiple metrics. On HumanML3D, LG-Tok achieves the best performance with a Top-1 R-Precision of \textbf{0.542} and FID of \textbf{0.057}, surpassing the previous best method, MoSa (0.518 and 0.064, respectively). 
On the larger and more diverse Motion-X dataset, our methods show substantial improvements, with LG-Tok-mid achieving the best Top-1 R-Precision of \textbf{0.591}. The consistent performance gains validate that transformer-based, language-guided tokenization effectively captures motion semantics. Remarkably, on both datasets, LG-Tok-mini maintains or surpasses state-of-the-art performance even after pruning more than half of the tokens (104 vs. 236), further validating the effectiveness of our high-level semantic representations. Results on the KIT-ML dataset are provided in Appendix~\ref{sec:kit_ml}.

\begin{figure*}[!t]
    \centering
    \includegraphics[width=1.0\textwidth, trim=1mm 0mm 5mm 0mm, clip]{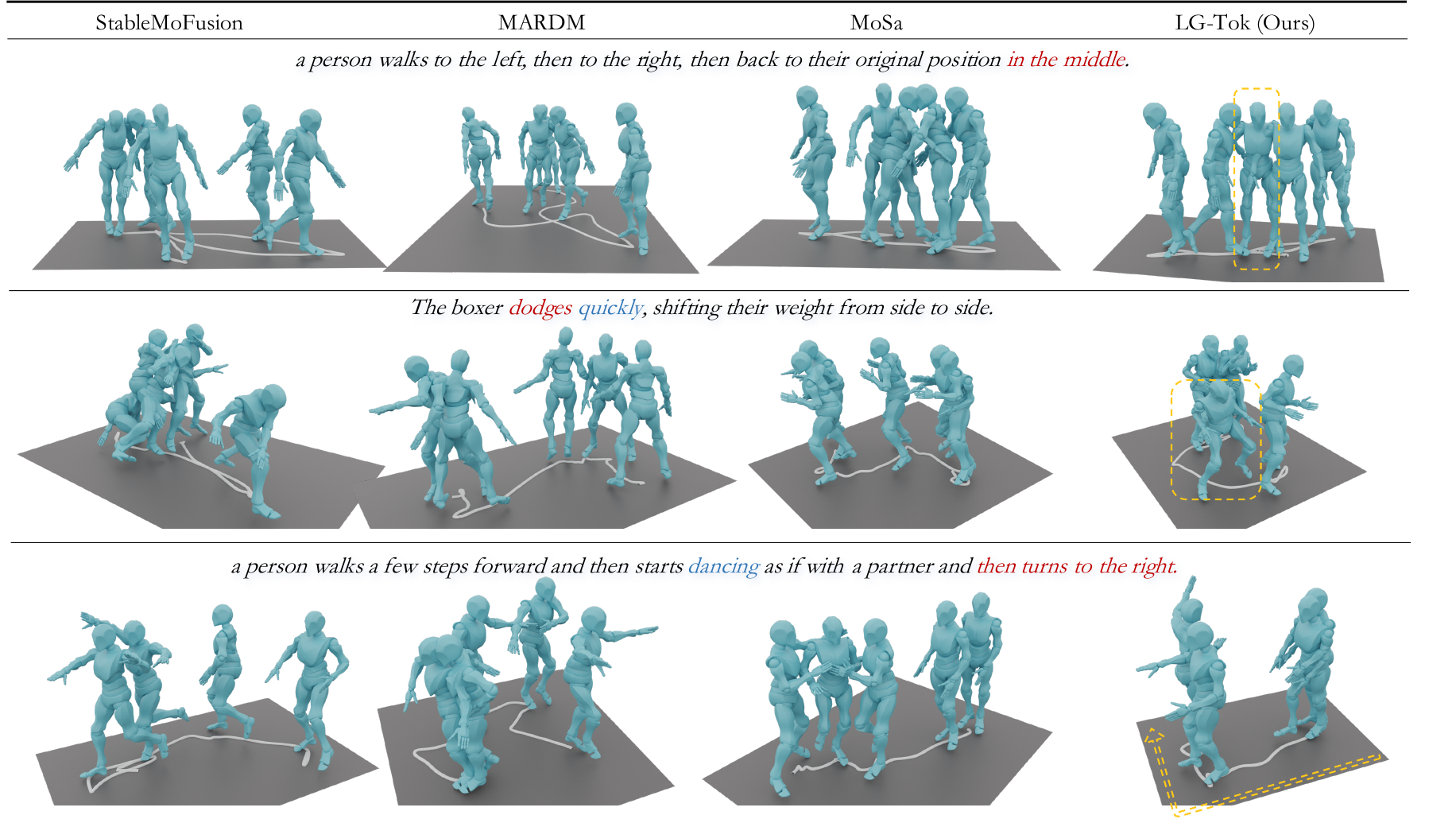}
    \vspace{-20pt}
    \caption{\textbf{Qualitative comparisons on HumanML3D dataset.} Our LG-Tok demonstrates superior semantic understanding compared to existing methods. The examples show better spatial awareness (``\textit{in the middle}"), more realistic posture synthesis (``\textit{dodges quickly}"), and improved directional control (``\textit{then turns to the left}").}
    \label{fig:comparasion}
\end{figure*}

\noindent\textbf{Qualitative comparisons.} 
Fig.~\ref{fig:comparasion} displays qualitative comparisons of our approach against StableMoFusion~\cite{huang2024stablemofusion}, MARDM~\cite{meng2025rethinking}, and MoSa~\cite{liu2025mosa} based on HumanML3D checkpoints. We randomly select several generated samples for comparison. These three examples demonstrate that LG-Tok can understand more complex semantics. \eg, in the first row, our method successfully understands ``\textit{in the middle}" (competitors fail to return to the middle position). In the second row, LG-Tok synthesizes more realistic crouching postures in the ``\textit{dodges quickly}" motion. In the final row, our approach exhibits better directional awareness for ``\textit{then turns to the left}". More dynamic results can be found on our supplementary materials.

\subsection{Comparison of Discrete Tokenizers}
\label{sec:vq_tok}

\input{table/table2}
We compare our approach against existing discrete (VQ-based) tokenizers, which are predominantly CNN-based. Table~\ref{tab:table2} demonstrates our LG-Tok consistent improvements in both reconstruction and generation stages across HumanML3D and Motion-X. Notably, by incorporating language guidance, our compact semantic representations significantly simplify the learning of generative models. On Motion-X, text-guided LG-Tok achieves substantial generation performance gains (FID: 0.257$\to$0.088), and the perplexity we recorded decreases from 146.5 to 103.1; similar improvements on HumanML3D (160.6$\to$155.9 ). These results confirm that language as auxiliary guidance relieves tokens from encoding high-level semantics, enabling focus on fine-grained motion details.

Furthermore, without text guidance (\ie, using only transformer-based tokenization from Sec.~\ref{subsec:transformer}), our method outperforms most competitors, demonstrating that our transformer-based architecture provides superior global context modeling. More importantly, we integrate LG-Tok into a representative motion tokenization–plus–generation framework, MoMask. The enhanced variant, MoMask (LG-Tok), consistently surpasses the original across all evaluation metrics, demonstrating strong generalization.

\subsection{In-depth Analysis of Language-Drop Scheme}
\label{sec:analysis_of_language_drop}

Table~\ref{tab:lang_drop_ablation} validates the language-drop scheme. A revealing pattern emerges consistently across all three datasets: without the language-drop scheme, reconstruction Top-1 and MM-Dist are marginally \textit{better}, yet reconstruction FID degrades substantially. This discrepancy suggests a shortcut effect: without language-drop, the tokenizer may over-rely on well-trained text embeddings, causing residual text priors to be embedded into the reconstructed motions, which could artificially inflate text-motion alignment scores in the joint embedding space. In contrast, FID---which directly measures the distributional distance between motion features---appears more sensitive to the degraded motion fidelity. Our language-drop scheme is designed to mitigate this shortcut by compelling both the tokenizer and detokenizer to operate without text, encouraging discrete tokens to be self-sufficient representations of motion, thereby restoring FID to healthy levels. The resulting tokens, which captured fine-grained motion details, also stabilize generation quality. This scheme further enables language-free guidance at inference: sweeping $g$ in Fig.~\ref{fig:ld} shows consistent gains across all datasets.

\input{table/table4_shortcut}

\subsection{Ablation Studies}
\label{sec:ablation}

\input{table/table3}

The remaining ablation studies were conducted on a smaller version (LG-Tok-tiny) with reduced model size and training data for efficient hyperparameter tuning. LG-Tok-tiny uses 3-layer transformers trained on 5,000 samples.

\noindent\textbf{Transformer architecture.} Tables~\ref{tab:ablation_normal}-~\ref{tab:ablation_pe} demonstrate the advantages of our transformer design over vanilla transformers~\cite{vaswani2017attention}. RMSNorm, SwiGLU activation, and skip connections all contribute to improved performance. Notably, RoPE with \textit{base}=100 is better suited for our short sequence task ($\sim$10s, 196 frames) compared to other bases.

\noindent\textbf{Guidance location.} Table~\ref{tab:cond_location} validates our method's benefits. Injecting guidance into both tokenizer and detokenizer achieves the best results, confirming that language guidance: 1) enables latent tokens to learn motion representations while simultaneously absorbing linguistic semantic knowledge during tokenizing; 2) allows language conditions to assist masked token sequences in effectively reconstructing input motion during detokenizing.

\noindent\textbf{Text encoder choice.} We compare popular pretrained text encoders~\cite{warner2024smarter,chung2024scaling,grattafiori2024llama,radford2021learning} in Table~\ref{tab:text_enocder}. LLaMA achieves the best performance among the tested encoders.

\noindent\textbf{Interaction methods.}  Table~\ref{tab:interaction_method} ablates how tokens interact in both the tokenizer and detokenizer. We compare two interaction strategies: \textit{In-context} (concatenating all tokens for joint self-attention) and \textit{Cross-attention} (two separate cross-attention between different token types). The results show that In-context concatenation performs better in the tokenizer, while cross-attention excels in the detokenizer.

\begin{figure}[!t]
  \centering
  \begin{subfigure}{0.49\linewidth}
    \centering
    \includegraphics[width=\linewidth]{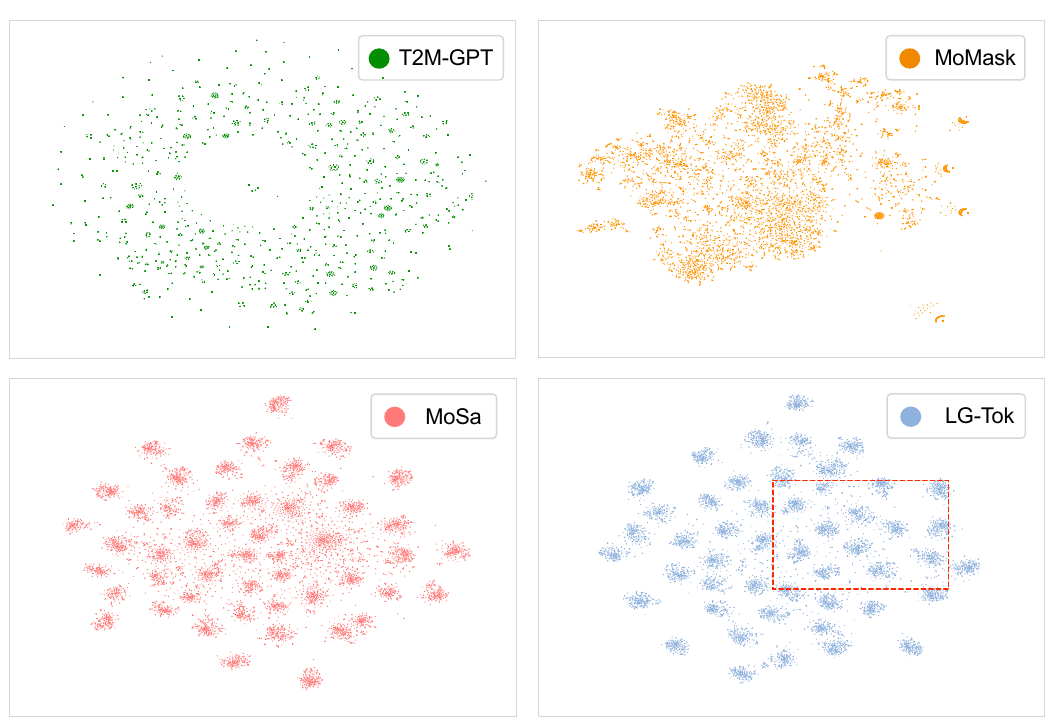}
  \end{subfigure}
  \hfill
  \begin{subfigure}{0.49\linewidth}
    \centering
    \includegraphics[width=\linewidth, trim=4mm 2mm 7.5mm 2mm, clip]{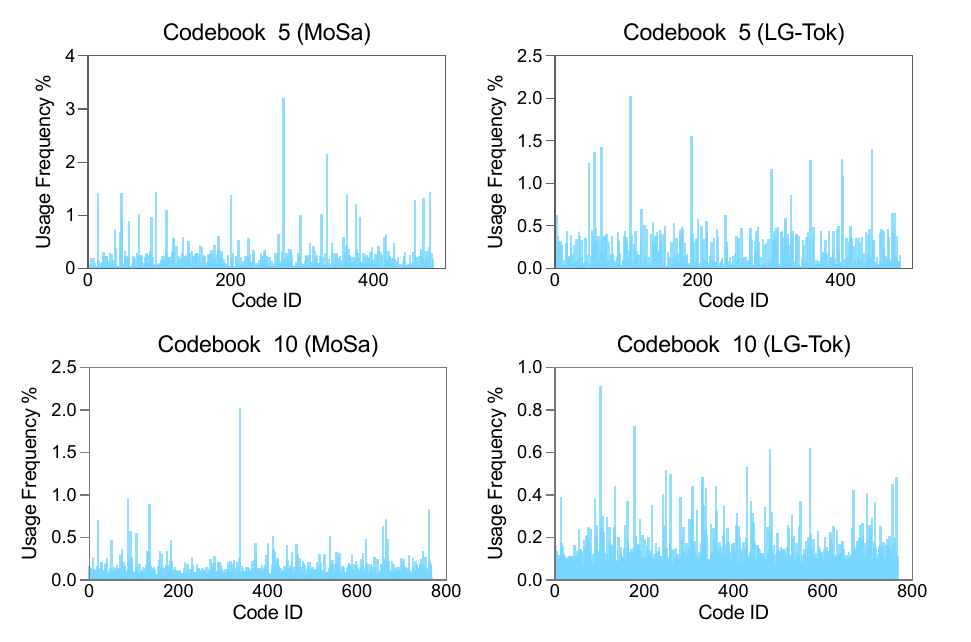}
  \end{subfigure}
  \vspace{-5pt}
  \caption{\textbf{Tokenizer analysis.} Left: t-SNE visualization of dequantized embedding space representations. LG-Tok demonstrates well-structured clusters with clear boundaries, indicating effective capture of distinct motion patterns. Right: code usage comparison on the HumanML3D test-set. LG-Tok achieves more uniform codebook utilization.}
  \label{fig:tokenizer_analysis}
  \vspace{-10pt}
\end{figure}

\subsection{Tokenizer Analysis}
\label{sec:tokenizer_analysis}

We provide further analysis of the learned representations of our tokenizer. As shown in Fig.~\ref{fig:tokenizer_analysis} (Left), the t-SNE representations of dequantized embeddings from different discrete tokenizers reveal that our LG-Tok demonstrates well-structured clusters with clear boundaries (\eg, in the red box), indicating that our compact, high-level semantic representations effectively capture distinct motion patterns. Fig.~\ref{fig:tokenizer_analysis} (Right) compares codebook usage frequencies across different scales on the HumanML3D test-set. LG-Tok achieves more uniform codebook utilization across all scales, indicating better coverage of the motion space. This uniform usage demonstrates that our language-guided, transformer-based approach possesses learned comprehensive motion representations without redundancy.

\subsection{Reconstruction--Generation Trade-off Analysis}
\label{sec:tradeoff}

\begin{wraptable}[12]{r}{0.6\columnwidth}
\centering
\vspace{-10pt}
\setlength{\tabcolsep}{2.5pt}
\renewcommand{\arraystretch}{1.0}
\caption{\textbf{Trade-off across LG-Tok variants.}}
\label{tab:tradeoff}
\resizebox{0.6\columnwidth}{!}{%
\begin{tabular}{@{}l|ccc|ccc@{}}
\toprule
 & \multicolumn{3}{c|}{HumanML3D} & \multicolumn{3}{c}{Motion-X} \\
\cmidrule(lr){2-4}\cmidrule(l){5-7}
 & LG-Tok-mini & LG-Tok-mid & LG-Tok & LG-Tok-mini & LG-Tok-mid & LG-Tok \\
\# Tokens & 104 & 160 & 236 & 104 & 160 & 236 \\
\hline
\rowcolor{cyan!6}\multicolumn{7}{c}{Reconstruction} \\ \hline
rFID$\downarrow$    & \et{0.094}{.001} & \et{0.052}{.000} & \et{\textbf{0.022}}{.000} & \et{0.094}{.002} & \et{0.051}{.001} & \et{\textbf{0.041}}{.001} \\
MPJPE$\downarrow$   & \et{50.2}{.0}    & \et{40.6}{.0}    & \et{\textbf{39.3}}{.0}    & \et{36.4}{.0}    & \et{34.2}{.0}    & \et{\textbf{31.7}}{.0}    \\
\hline
\rowcolor{cyan!6}\multicolumn{7}{c}{Generation} \\ \hline
gFID$\downarrow$    & \et{0.085}{.004} & \et{0.109}{.005} & \et{\textbf{0.057}}{.003} & \et{\textbf{0.071}}{.004} & \et{0.076}{.005} & \et{0.088}{.006} \\
Top-1$\uparrow$     & \et{0.521}{.003} & \et{0.537}{.003} & \et{\textbf{0.542}}{.003} & \et{0.588}{.002}           & \et{\textbf{0.591}}{.002} & \et{0.582}{.002} \\
MM-Dist$\downarrow$ & \et{3.113}{.010} & \et{3.061}{.010} & \et{\textbf{2.997}}{.010} & \et{3.835}{.012}           & \et{\textbf{3.758}}{.010} & \et{3.844}{.009} \\
CLIP$\uparrow$      & \et{0.655}{.001} & \et{0.664}{.001} & \et{\textbf{0.669}}{.001} & \et{0.681}{.001}           & \et{\textbf{0.682}}{.000} & \et{\textbf{0.682}}{.000} \\
\bottomrule
\end{tabular}}
\end{wraptable}
Table~\ref{tab:tradeoff} makes concrete the trade-off raised in Sec.~\ref{sec:intro}, where generation quality reflects two competing forces: more tokens raise the representational ceiling and benefit reconstruction, while longer sequences burden the autoregressive model. Which force prevails may hinge on the amount of training data, we think. On the smaller HumanML3D (14k sequences), the nearly fourfold drop in reconstruction error (rFID from 0.094 to 0.022) lets the former dominate, and the full LG-Tok with 236 tokens wins on every metric. On the larger Motion-X (37k), the gain is far milder (roughly twofold, rFID from 0.094 to 0.041), and beyond 160 tokens the latter takes over: generation degrades (gFID from 0.071 to 0.088), leaving LG-Tok-mid as the sweet spot.

%% file: table/table1.tex
\begin{table*}[t]
\centering
\caption{\textbf{Quantitative evaluation on the 
HumanML3D and Motion-X test set.} Each experiment is
evaluated 20 times, and $\pm$ indicates a 95\% confidence interval. \textbf{Bold} and \underline{underline} indicate the best and the second best result, respectively. `$\dagger$' denotes our reimplementation on the \texttt{meng} representation~\cite{meng2025rethinking}.}
\label{tab:table1} \vspace{-5pt}
\renewcommand{\arraystretch}{1.2}
\resizebox{1.0\textwidth}{!}{%
\begin{tabular}{@{}lllcccccccc@{}}
\toprule
\multirow{2}{*}{Datasets} & \multirow{2}{*}{Methods} & \multirow{2}{*}{Venues} & \multirow{2}{*}{\#Tokens} & \multicolumn{3}{c}{R Precision$\uparrow$} & \multirow{2}{*}{FID$\downarrow$} & \multirow{2}{*}{MultiModal Dist$\downarrow$} & \multirow{2}{*}{MultiModality$\uparrow$} & \multirow{2}{*}{CLIP-score$\uparrow$} \\ \cmidrule(lr){5-7}
 &  &  &  & Top 1 & Top 2 & Top 3 &  &  &  &  \\ \midrule
\multirow{13}{*}{\makecell[c]{Human\\ML3D}} & \textbf{Real motions} & - & - & \et{0.501}{.002} & \et{0.696}{.003} & \et{0.792}{.002} & \et{0.000}{.000} & \et{3.251}{.010} & - & \et{0.639}{.001} \\ \cmidrule(l){2-11} 
 & MotionDiffuse~\cite{zhang2022motiondiffuse} & TPAMI\textquotesingle\,24 & - & \et{0.450}{.006} & \et{0.641}{.005} & \et{0.753}{.005} & \et{0.778}{.005} & \et{3.490}{.023} & \etb{3.179}{.046} & \et{0.606}{.004} \\
 & T2M-GPT~\cite{zhang2023generating} & CVPR\textquotesingle\,23 & 49 & \et{0.470}{.003} & \et{0.659}{.002} & \et{0.758}{.002} & \et{0.335}{.003} & \et{3.505}{.017} & \et{2.018}{.053} & \et{0.607}{.005} \\
 & MMM~\cite{pinyoanuntapong2024mmm} & CVPR\textquotesingle\,24 & 49 & \et{0.487}{.003} & \et{0.683}{.003} & \et{0.782}{.001} & \et{0.132}{.004} & \et{3.359}{.009} & \et{1.241}{.073} & \et{0.635}{.003} \\
 & MoMask~\cite{guo2024momask} & CVPR\textquotesingle\,24 & 294 & \et{0.490}{.004} & \et{0.687}{.003} & \et{0.786}{.003} & \et{0.116}{.006} & \et{3.353}{.010} & \et{1.263}{.079} & \et{0.637}{.003} \\
 & StableMoFusion$^{\dagger}$~\cite{huang2024stablemofusion} & ACM MM\textquotesingle\,24 & - & \et{0.510}{.002} & \et{0.710}{.003} & \et{0.810}{.003} & \et{0.177}{.006} & \et{3.182}{.009} & \et{1.969}{.053} & \et{0.654}{.001}  \\
 & DisCoRD$^{\dagger}$~\cite{cho2025discord} & ICCV\textquotesingle\,25 & - & \et{0.492}{.003} & \et{0.690}{.003} & \et{0.790}{.003} & \et{0.117}{.004} & \et{3.306}{.008} & \et{1.369}{.061} & \et{0.640}{.001} \\
 & MARDM~\cite{meng2025rethinking} & CVPR\textquotesingle\,25 & - & \et{0.500}{.004} & \et{0.695}{.003} & \et{0.795}{.003} & \et{0.114}{.007} & \et{3.270}{.009} & \ets{2.231}{.071} & \et{0.642}{.002} \\
 & MoSa$^{\dagger}$~\cite{liu2025mosa} & Arxiv\textquotesingle\,25 & 236 & \et{0.518}{.002} & \et{0.712}{.002} & \et{0.809}{.002} & \ets{0.064}{.004} & \et{3.150}{.008} & \et{1.789}{.056} & \et{0.657}{.001} \\ \cmidrule(l){2-11}
 & \textbf{LG-Tok-mini} & - & 104 & \et{0.521}{.003} & \et{0.715}{.003}  & \et{0.811}{.003} & \et{0.085}{.004} & \et{3.113}{.001} & \et{1.728}{.053} & \et{0.655}{.001} \\
 & \textbf{LG-Tok-mid} & - & 160 & \ets{0.537}{.003} & \ets{0.729}{.002} & \ets{0.821}{.002} & \et{0.109}{.005} & \ets{3.061}{.010} & \et{1.674}{.052} & \ets{0.664}{.001} \\
 & \textbf{LG-Tok} & - & 236 & \etb{0.542}{.003} & \etb{0.736}{.002} & \etb{0.830}{.002} & \etb{0.057}{.003} & \etb{2.997}{.010} & \et{1.540}{.059} & \etb{0.669}{.001} \\ \midrule
\multirow{13}{*}{\makecell[c]{Motion-\\X}} & \textbf{Real motions} & - & - & \et{0.595}{.002} & \et{0.787}{.001} & \et{0.868}{.001} & \et{0.000}{.000} & \et{3.717}{.007} & - & \et{0.672}{.000} \\ \cmidrule(l){2-11} 
 & MotionDiffuse$^{\dagger}$~\cite{zhang2022motiondiffuse} & TPAMI\textquotesingle\,24 & - & \et{0.559}{.003} & \et{0.752}{.002} & \et{0.839}{.001} & \et{0.954}{.014} & \et{4.167}{.023} & \et{2.476}{.098} & \et{0.660}{.001} \\
 & T2M-GPT$^{\dagger}$~\cite{zhang2023generating} & CVPR\textquotesingle\,23 & 49 & \et{0.470}{.003} & \et{0.644}{.002} & \et{0.735}{.003} & \et{1.085}{.032} & \et{5.488}{.023} & \etb{12.807}{.405} &  \et{0.622}{.001} \\
 & MMM$^{\dagger}$~\cite{pinyoanuntapong2024mmm} & CVPR\textquotesingle\,24 & 49 & \et{0.424}{.002} & \et{0.600}{.002} & \et{0.695}{.002} & \et{2.918}{.036} & \et{6.098}{.019} & \et{2.342}{.193} & \et{0.607}{.001} \\
 & MoMask$^{\dagger}$~\cite{guo2024momask} & CVPR\textquotesingle\,24 & 294 & \et{0.502}{.003} & \et{0.694}{.002} & \et{0.790}{.002} & \et{0.247}{.008} & \et{4.832}{.009} & \et{2.715}{.091} & \et{0.644}{.001} \\
 & StableMoFusion$^{\dagger}$~\cite{huang2024stablemofusion} & ACM MM\textquotesingle\,24 & - & \et{0.474}{.003} & \et{0.682}{.002} & \et{0.787}{.002} & \et{0.213}{.008} & \et{4.888}{.014} & \et{2.985}{.111} & \et{0.607}{.001} \\
 & DisCoRD$^{\dagger}$~\cite{cho2025discord} & ICCV\textquotesingle\,25 & - & \et{0.518}{.002} & \et{0.714}{.002} & \et{0.808}{.002} & \et{0.164}{.008} & \et{4.614}{.014} & \et{2.417}{.112} & \et{0.649}{.000} \\
 & MARDM$^{\dagger}$~\cite{meng2025rethinking} & CVPR\textquotesingle\,25 & - & \et{0.528}{.003} & \et{0.727}{.001} & \et{0.820}{.002} & \et{0.147}{.009} & \et{4.433}{.018} & \et{3.077}{.066} & \et{0.643}{.001} \\
 & MoSa$^{\dagger}$~\cite{liu2025mosa} & Arxiv\textquotesingle\,25 & 236 & \et{0.513}{.002} & \et{0.703}{.002} & \et{0.796}{.002} & \et{0.210}{.009} & \et{4.783}{.013} & \ets{3.167}{.122} & \et{0.654}{.001} \\ \cmidrule(l){2-11}
 & \textbf{LG-Tok-mini} & - & 104 & \ets{0.588}{.002} & \ets{0.776}{.002} & \et{0.857}{.002} & \etb{0.071}{.004} & \ets{3.835}{.012} & \et{2.235}{.111} & \ets{0.681}{.001} \\
 & \textbf{LG-Tok-mid} & - & 160 & \etb{0.591}{.002} & \etb{0.781}{.002} & \etb{0.864}{.001} & \ets{0.076}{.005} & \etb{3.758}{.010} & \et{2.245}{.091} & \etb{0.682}{.000} \\
 & \textbf{LG-Tok} & - & 236 & \et{0.582}{.002} & \et{0.775}{.001} & \ets{0.858}{.001} & \et{0.088}{.006} & \et{3.844}{.009} & \et{2.294}{.088} & \etb{0.682}{.000} \\ \bottomrule
\end{tabular}%
}
\end{table*}

%% file: table/table2.tex
\begin{wraptable}[23]{r}{0.55\textwidth}
\vspace{-10pt}
\caption{\textbf{Reconstruction and generation performance comparison.} We evaluate both reconstruction quality and generation performance across different discrete tokenizers. ``\textit{w/o} text guidance" denotes our approach without language guidance, \ie, using only the transformer-based tokenization from Sec.~\ref{subsec:transformer}. ``$\star$" denotes report from~\cite{meng2025rethinking}, and the others are our reproductions.}
\label{tab:table2}
\centering
\tabcolsep=2pt
\renewcommand{\arraystretch}{1.3}
\resizebox{\linewidth}{!}{%
\begin{tabular}{@{}l|cccc|cc@{}}
\toprule 
\multirow{2}{*}{Methods} & \multicolumn{3}{c}{Reconstruction} &  & \multicolumn{2}{c}{Generation} \\ \cmidrule(lr){2-4} \cmidrule(l){6-7} 
 & FID$\downarrow$ & Top 1$\uparrow$ & MPJPE$\downarrow$ &  & FID$\downarrow$ & MM-Dist$\downarrow$ \\ \hline
\bluerow
\multicolumn{7}{c}{Evaluation on HumanML3D dataset} \\ \hline
T2M-GPT{\large $^\star$}~\cite{zhang2023generating} & \et{0.081}{.001} & \et{0.483}{.003} & \et{72.6}{.001} &  & \et{0.335}{.003} & \et{3.505}{.017} \\
MoMask{\large $^\star$}~\cite{guo2024momask} & \et{0.029}{.001} & \et{0.497}{.002} & \et{31.5}{.001} &  & \et{0.116}{.006} & \et{3.353}{.010} \\
MoMask-reprod. & \et{0.029}{.000} & \et{0.499}{.003} & \et{30.9}{.000} &  & \et{0.180}{.005} & \et{3.332}{.009} \\
MoMask (LG-Tok) & \etb{0.019}{.000} & \et{0.501}{.003} & \etb{26.4}{.000} &  & \et{0.111}{.005} & \et{3.300}{.012} \\
MoSa~\cite{liu2025mosa} & \et{0.023}{.000} & \et{0.496}{.003} & \et{43.0}{.000} &  & \et{0.064}{.004} & \et{3.150}{.008} \\
LG-Tok & \et{0.022}{.000} & \etb{0.502}{.002} & \et{39.0}{.000} &  & \etb{0.057}{.003} & \etb{2.997}{.010} \\ 
\,\, \textit{w/o} text guidance\,  & \et{0.025}{.000} & \et{0.494}{.002} & \et{39.0}{.000} &  & \et{0.062}{.003} & \et{3.129}{.008} \\ \hline
\bluerow
\multicolumn{7}{c}{Evaluation on Motion-X dataset} \\ \hline
T2M-GPT~\cite{zhang2023generating} & \et{0.826}{.011} & \et{0.497}{.002} & \et{59.6}{.000} &  & \et{1.102}{.000} & \et{5.515}{.000} \\
MoMask~\cite{guo2024momask} & \et{0.394}{.005} & \et{0.554}{.002} & \et{24.9}{.000} &  & \et{0.247}{.008} & \et{4.832}{.009} \\
MoMask (LG-Tok) & \et{0.076}{.002} & \etb{0.580}{.002} & \etb{23.0}{.000} &  & \et{0.157}{.006} & \et{4.259}{.008}  \\
MoSa~\cite{liu2025mosa} & \et{0.072}{.001} & \et{0.558}{.002} & \et{39.0}{.000} &  & \et{0.210}{.009} & \et{4.783}{.013} \\
LG-Tok & \etb{0.041}{.001} & \et{0.577}{.002} & \et{31.0}{.000} &  & \etb{0.088}{.006} & \etb{3.844}{.009} \\
\,\, \textit{w/o} text guidance\, & \et{0.090}{.002} & \et{0.568}{.002} & \et{33.0}{.000} &  & \et{0.257}{.010} & \et{4.274}{.008} \\ \bottomrule
\end{tabular}
}
\end{wraptable}

%% file: table/table4_shortcut.tex
\begin{figure*}[!ht]
\centering
\vspace{-10pt}
\begin{minipage}{0.64\textwidth}
\begin{table}[H]
\renewcommand{\arraystretch}{1.4}
\centering
\resizebox{\textwidth}{!}{%
\begin{tabular}{l|ccc|ccc}
\toprule
\multirow{2}{*}{\textbf{Method}} & \multicolumn{3}{c|}{\textbf{Reconstruction}} & \multicolumn{3}{c}{\textbf{Generation}} \\
 & \multicolumn{1}{c}{FID$\downarrow$} & \multicolumn{1}{c}{Top 1$\uparrow$} & \multicolumn{1}{c|}{MM-Dist.$\downarrow$} & \multicolumn{1}{c}{FID$\downarrow$} & \multicolumn{1}{c}{Top 1$\uparrow$} & \multicolumn{1}{c}{MM-Dist.$\downarrow$} \\ \hline
\bluerow
\multicolumn{7}{c}{Evaluation on HumanML3D dataset} \\ \hline
\rowcolor{gray!8} LG-Tok ($g=2.0$) &  &  &  & \etb{0.057}{.003} & \etb{0.542}{.003} & \etb{2.997}{.010} \\
\rowcolor{gray!8} LG-Tok ($g=0.0$) & \multirow{-2}{*}{\etb{0.022}{.000}} & \multirow{-2}{*}{\et{0.502}{.002}} & \multirow{-2}{*}{\et{3.250}{.010}} & \et{0.061}{.003} & \et{0.534}{.003} & \et{3.032}{.011} \\
\,\, \textit{w/o} lang-drop scheme\,  & \et{0.039}{.000} & \etb{0.505}{.002} & \etb{3.233}{.009} & \et{0.132}{.005} & \et{0.533}{.003} & \et{3.068}{.007} \\
\hline \bluerow
\multicolumn{7}{c}{Evaluation on Motion-X dataset} \\ \hline
\rowcolor{gray!8} LG-Tok ($g=1.0$) &  &  &  & \etb{0.088}{.006} & \etb{0.582}{.002} & \etb{3.844}{.009} \\
\rowcolor{gray!8} LG-Tok ($g=0.0$) & \multirow{-2}{*}{\etb{0.041}{.001}} & \multirow{-2}{*}{\et{0.577}{.002}} & \multirow{-2}{*}{\et{3.890}{.008}} & \et{0.139}{.008} & \et{0.576}{.002} & \et{3.969}{.012} \\
\,\, \textit{w/o} lang-drop scheme\, & \et{0.057}{.002} & \etb{0.580}{.002} & \etb{3.846}{.007} & \et{0.136}{.007} & \et{0.574}{.002} & \et{3.918}{.012} \\
\hline \bluerow
\multicolumn{7}{c}{Evaluation on KIT-ML dataset} \\ \hline
\rowcolor{gray!8} LG-Tok ($g=2.0$) &  &  &  & \etb{0.185}{.008} & \etb{0.401}{.006} & \etb{3.270}{.013} \\
\rowcolor{gray!8} LG-Tok ($g=0.0$) & \multirow{-2}{*}{\etb{0.108}{.003}} & \multirow{-2}{*}{\et{0.373}{.005}} & \multirow{-2}{*}{\et{3.395}{.011}} & \et{0.241}{.014} & \et{0.390}{.006} & \et{3.331}{.017} \\
\,\, \textit{w/o} lang-drop scheme\, & \et{0.139}{.004} & \etb{0.378}{.006} & \etb{3.324}{.011} & \et{0.251}{.015} & \et{0.375}{.005} & \et{3.469}{.013} \\
\bottomrule
\end{tabular}%
}
\vspace{5pt}
\caption{\textbf{In-depth analysis of language-drop scheme.} Symbol $g$ denotes the guidance scale for language-free decoding during generation. ``\textit{w/o} lang-drop scheme'' denotes the tokenizer trained with language guidance but without randomly dropping text. }
\label{tab:lang_drop_ablation}
\end{table}
\end{minipage}%
\hfill
\nextfloat
\begin{minipage}{0.34\textwidth}
  \centering
  \vspace{18pt}
  \subfloat[HumanML3D]{%
    \includegraphics[width=0.795\linewidth, trim=2mm 0mm 2mm 2mm, clip]{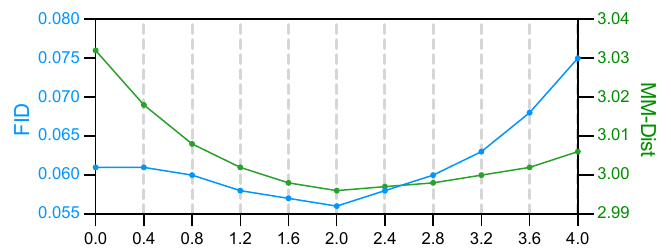}%
    \label{fig:ld-humanml3d}%
  }
  \\
  \vspace{0mm}
  \subfloat[Motion-X]{%
    \includegraphics[width=0.8\linewidth, trim=0mm 0mm 2mm 2mm, clip]{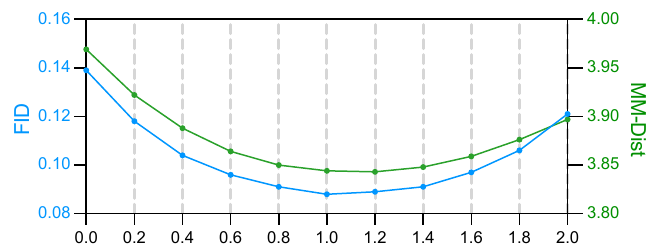}%
    \label{fig:ld-motionx}%
  }
  \\
  \vspace{0mm}
  \subfloat[KIT-ML]{%
    \includegraphics[width=0.8\linewidth, trim=0mm 0mm 2mm 2mm, clip]{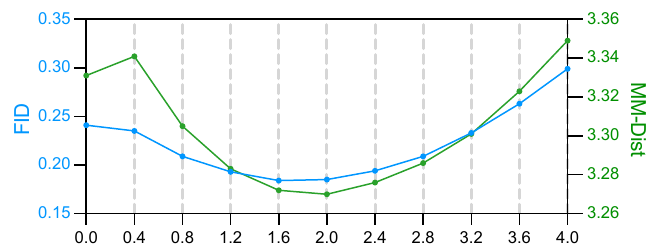}%
    \label{fig:ld-kit-ml}%
  }
  \vspace{-8pt}
  \caption{\textbf{Evaluation sweep over guidance scale $g$.} $g=2.0$, $g=1.0$ and $g=2.0$ yield the best performance for HumanML3D, Motion-X and KIT-ML, respectively.}
  \label{fig:ld}
\end{minipage}
\vspace{-15pt}
\end{figure*}

%% file: table/table3.tex
\begin{table*}[t]
\caption{\textbf{Ablation studies on LG-Tok-tiny}. We ablate key design choices affecting LG-Tok's reconstruction performance on HumanML3D dataset. \colorbox{baselinecolor}{Default setting:} RMSNorm, SwiGLU activation, skip connections, RoPE with \textit{base}=100, LLaMA text encoder, with in-context conditioning for tokenizer and cross-attention for detokenizer. Injecting natural language to both tokenizer and detokenizer obtains the best results.} \vspace{-4pt}
\label{tab:ablations}
\centering
\tiny
\subfloat[
\scriptsize Normalization
\label{tab:ablation_normal}
]{
\centering
\begin{minipage}{0.3\linewidth}{\begin{center}
\resizebox{\linewidth}{!}{%
\begin{tabular}{y{46}x{36}x{24}} \toprule
Normalization & FID$\downarrow$ & MPJPE$\downarrow$ \\
\midrule
LayerNorm & \et{0.050}{.001} & 58.7 \\
\bluerow
RMSNorm & \et{0.049}{.000} & 56.1   \\
\end{tabular}
}
\end{center}}
\end{minipage}
}
\hspace{1em}
\subfloat[
\scriptsize Activation
\label{tab:ablation_act}
]{
\centering
\begin{minipage}{0.3\linewidth}{\begin{center}
\resizebox{\linewidth}{!}{%
\begin{tabular}{y{46}x{36}x{24}} \toprule
Architecture & FID$\downarrow$ & MPJPE$\downarrow$ \\
\midrule
GeLU & \et{0.050}{.001} & 57.4 \\
\bluerow
SwiGLU & \et{0.049}{.000} & 56.1   \\
\end{tabular}
}
\end{center}}
\end{minipage}
}
\hspace{1em}
\subfloat[
\scriptsize Skip connections
\label{tab:ablation_skip}
]{
\centering
\begin{minipage}{0.3\linewidth}{
\begin{center}
\resizebox{\linewidth}{!}{%
\begin{tabular}{y{50}x{36}x{24}} \toprule
Activation  & FID$\downarrow$ & MPJPE$\downarrow$ \\
\midrule
\bluerow
Skip Conn. & \et{0.049}{.000} & 56.1    \\
\textit{w/o} Skip Conn. & \et{0.059}{.001}  &  61.7 \\
\end{tabular}
}
\end{center}}
\end{minipage}
}
\\
\vspace{1mm}
\subfloat[
\scriptsize Position embedding
\label{tab:ablation_pe}
]{
\begin{minipage}{0.44\linewidth}{\begin{center}
\resizebox{\linewidth}{!}{%
\begin{tabular}{y{66}x{36}x{24}} \toprule
Position embedding & FID$\downarrow$ & MPJPE$\downarrow$ \\
\midrule
Learnable & \et{0.065}{.001} & 54.7 \\ 
RoPE (\textit{base}=10) & \et{0.061}{.001} & 54.8 \\
\bluerow
RoPE (\textit{base}=100) & \et{0.049}{.000} & 56.1   \\
RoPE (\textit{base}=1000) & \et{0.042}{.001} & 56.3 \\
RoPE (\textit{base}=10000) & \et{0.052}{.001} & 56.6 \\
\end{tabular}
}
\end{center}}\vspace{-4mm}\end{minipage}
}
\hspace{1em}
\subfloat[
\scriptsize Guidance location
\label{tab:cond_location}
]{
\begin{minipage}{0.44\linewidth}{\begin{center}
\resizebox{\linewidth}{!}{%
\begin{tabular}{x{76}x{36}x{24}} \toprule
Guidance location & FID$\downarrow$ & MPJPE$\downarrow$ \\
\midrule
None & \et{0.064}{.001} & 64.1 \\
Tokenizer only & \et{0.063}{.001} & 57.5 \\
Detokenizer only & \et{0.055}{.000} & 58.7 \\
\bluerow
Tokenizer \& Detokenizer & \et{0.049}{.000} & 56.1   \\
\multicolumn{3}{c}{~}\\
\end{tabular}
}
\end{center}}\vspace{-3.5mm}\end{minipage}
}
\\
\vspace{1mm}
\subfloat[
\scriptsize Text encoder chosen
\label{tab:text_enocder}
]{
\begin{minipage}{0.35\linewidth}{\begin{center}
\resizebox{\linewidth}{!}{%
\begin{tabular}{y{40}x{36}x{24}} \toprule
Frozen & \multirow{2}{*}{FID$\downarrow$} & \multirow{2}{*}{MPJPE$\downarrow$} \\
text encoder &  &  \\ 
\midrule
CLIP & \et{0.051}{.000} & 59.1 \\
T5 & \et{0.053}{.000} & 58.9 \\
BERT & \et{0.055}{.001} & 58.9 \\
\bluerow
LLaMA & \et{0.049}{.000} & 56.1  \\
\end{tabular}
}
\end{center}}\end{minipage}
}
\hspace{-0.5em}
\subfloat[
\scriptsize Latent-tokens/motion/text interaction in tokenizer \& Mask-tokens/latent/text interation in detokenizer \label{tab:interaction_method}
]{
\begin{minipage}{0.60\linewidth}{\begin{center}
\resizebox{\linewidth}{!}{%
\begin{tabular}{x{36}x{36}x{1}x{36}x{36}x{36}x{24}} \toprule
\multicolumn{2}{c}{Tokenizer} &  & \multicolumn{2}{c}{Detokenizer} & \multirow{2}{*}{FID$\downarrow$} & \multirow{2}{*}{MPJPE$\downarrow$} \\ \cmidrule(r){1-2} \cmidrule(lr){4-5}
In-Context & Cross-Attn. &  & In-Context & Cross-Attn. &  &  \\ \midrule
\ding{51} & \ding{55} &  & \ding{51} & \ding{55} & \et{0.053}{.001} & 66.1 \\
\ding{55}& \ding{51} &  & \ding{55} & \ding{51} & \et{1.120}{.004} & 111.2\\ 
\bluerow
\ding{51} & \ding{55} &  & \ding{55} & \ding{51} & \et{0.049}{.000} & 56.1 \\
\ding{55} & \ding{51} &  & \ding{51} & \ding{55} & \et{2.049}{.007} & 127.7 \\
\end{tabular}%
}
\end{center}}\end{minipage}
}
\end{table*}

%% file: sec/5_conclusion.tex
\section{Conclusion}

In this paper, we present Language-Guided Transformer Tokenization (LG-Tok) for text-driven human motion generation. By aligning natural language with motion during tokenization and employing a Transformer-based tokenizer-detokenizer architecture, LG-Tok provides semantically informed latent representations that improve reconstruction fidelity and simplify generative learning. LG-Tok achieves state-of-the-art results on three benchmarks.

%% file: sec/6_acknowledge.tex
\section*{Acknowledgements}
This work was supported by National Natural Science Foundation of China (No. 62473007), Guangdong Outstanding Youth Fund (No. 2026B1515020015), Shenzhen Innovation in Science and Technology Foundation for The Excellent Youth Scholars (No. RCYX20231211090248064).

%% file: sec/X_suppl.tex
\clearpage
\setcounter{page}{1}
\onecolumn
\appendix

\begin{center} 
    \centering
    \textbf{\Large Language-Guided Transformer Tokenizer for Human Motion Generation}
\end{center}
\begin{center} 
    \centering
    \large Supplementary Material
\end{center}

\section{Overview}
\label{sec:Summary}
The supplementary material is organized into the following sections:
\begin{itemize}
    \item Section \ref{sec:sub_tok_gen_detok_pipeline}: Complete tokenization-generation-detokenization pipeline
    \item Section \ref{sec:kit_ml}: Quantitative evaluation on the KIT-ML dataset
    \item Section \ref{sec:feature_extractor_training}: Feature extractor training on the Motion-X dataset
    \item Section \ref{sec:baseline_reprod}: Details of baseline reproduction on the Motion-X dataset
    \item Section \ref{sec:motion_editing}: Application: motion editing
    \item Section \ref{sec:average_inference_time}: Evaluation of average inference time
    \item Section \ref{sec:qualitative_comparsion_for_text_guidance}: Qualitative comparison for text guidance
    \item Section \ref{sec:role_of_language}: The role of language guidance in the detokenization stage during generation
    \item Section \ref{sec:more_qualitative_comparison}: More qualitative results
    \item Section \ref{sec:limitations}: Limitations
\end{itemize}

\section{Complete Tokenization-Generation-Detokenization Pipeline}
\label{sec:sub_tok_gen_detok_pipeline}

\begin{figure*}[!ht]
    \centering
    \includegraphics[width=1.0\textwidth, trim=0mm 0mm 0mm 0mm, clip]{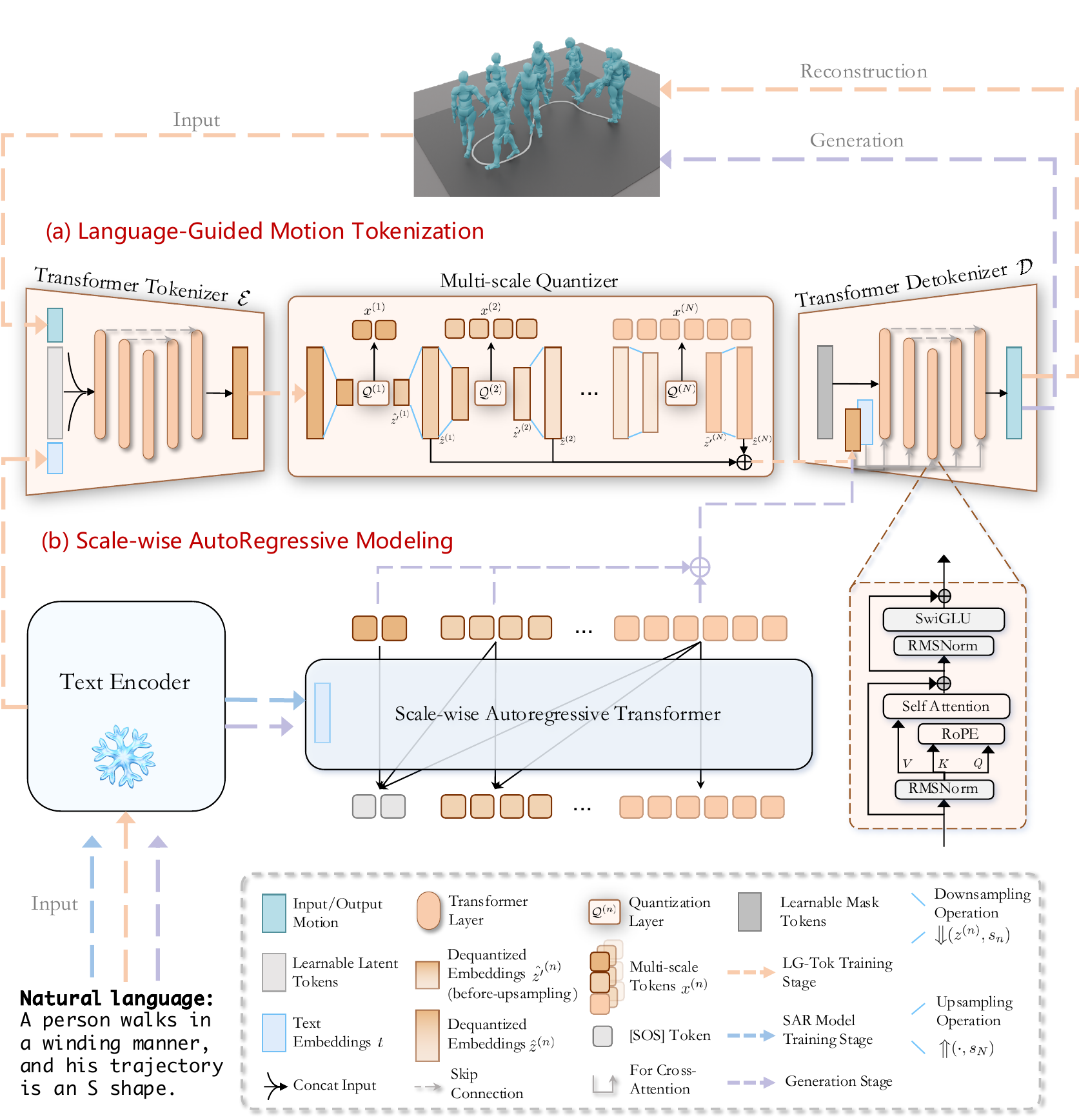} 
    \caption{\textbf{Complete tokenization-generation-detokenization pipeline.} (a) \textit{Language-Guided Motion Tokenization}: The transformer tokenizer encodes motion with text guidance, followed by multi-scale quantization that produces multi-scale token sets to enable (b) \textit{Scale-wise Autoregressive Modeling}: The SAR model performs scale-by-scale generation conditioned on text, and the transformer detokenizer decodes motion from dequantized embeddings. Orange, blue, and purple dashed lines indicate LG-Tok training, SAR Model training, and generation flows, respectively.}
    \label{fig:whole_pipeline}
\end{figure*}

In this section, we present the complete tokenization-generation-detokenization pipeline to facilitate a comprehensive understanding of the interplay between tokenization and generation. We adopt the Scale-wise AutoRegressive (SAR) modeling framework, MoSa~\cite{liu2025mosa}, as our generative model, benefiting from its notable performance gains. As illustrated in Fig.~\ref{fig:whole_pipeline}, we apply multi-scale quantization in (a) \textit{Language-Guided Motion Tokenization}, where the retained multi-scale token sets enable (b) \textit{Scale-wise AutoRegressive (SAR) Modeling}. In the SAR section, we demonstrate the training procedure. Finally, we illustrate how the trained LG-Tok and SAR models interact to generate high-quality, diverse motions conditioned on natural language descriptions. We elaborate on this pipeline through three stages:

\noindent\textbf{1) LG-Tok Training Stage.} The orange dashed line illustrates the tokenization flow. Since the tokenization process is detailed in Sec.~\ref{sec:method}, we focus on constructing the multi-scale quantizer. Unlike standard residual quantization, the multi-scale quantizer introduces a downsample-quantize-upsample interpolation at each residual quantization layer, enabling the model to obtain token representations at different temporal resolutions, all of which are retained for SAR modeling in stage 2). Mathematically, the resulting multi-scale token set is denoted as:
\begin{align}
    x = \left\{\underbrace{(x^{(1)}_1, \dots, x^{(1)}_{s_{1}})}_{x^{(1)}}, \underbrace{(x^{(2)}_1, x^{(2)}_2, \dots, x^{(2)}_{s_{2}})}_{x^{(2)}}, \dots, \underbrace{(x^{(N)}_1, x^{(N)}_2, \dots, x^{(N)}_{s_{N}})}_{x^{(N)}} \right\} \label{eq:multi-scale}
\end{align}  
where $x^{(n)} \in \mathbb{R}^{s_n}$ denotes the token sequence at the $n$-th scale, with length controlled by a predefined scale scheduler $S=(s_1,\dots,s_N)$. For instance, in LG-Tok-mid, $S=(2,4,\dots,36)$ increases progressively from the coarsest scale (2 tokens) to the finest scale (36 tokens). 

 We now describe how multi-scale tokens are obtained through the quantization process. After feature encoding by the tokenizer, the latent features $z\in \mathbb{R}^{T\times d}$ sequentially pass through $N$ sequential quantizer layers $\mathcal{Q}$. At each layer, A downsampling operation $\Downarrow(\cdot, s_n)$ is first applied to reduce the latent features $z^{(n)}$ to the target scale $s_n$. The downsampled features are then quantized by retrieving their nearest codebook entries, generating the $n$-th scale tokens (discrete codes) $x^{(n)} =\mathcal{Q}^{(n)}({\Downarrow}(z^{(n)}, s_n)) \in \mathbb{R}^{s_n}$. The corresponding codebook entries form the dequantized embeddings at this scale $\hat{z'}^{(n)}\in \mathbb{R}^{{s_n}\times d}$. To enable proper residual updates across different scales, the dequantized embeddings $\hat{z'}^{(n)}$ must be restored to a consistent scale. An upsampling operation  $\Uparrow(\cdot)$  is applied to bring the embedding scale back to $s_N$: $\hat{z}^{(n)}=\Uparrow(\hat{z'}^{(n)}) \in \mathbb{R}^{s_{N} \times d}$. This allows the computation of the residual for the next quantization layer: $z^{(n+1)}=z^{(n)}-\hat{z}^{(n)}$. 
 
 The downsample-quantize-upsample process is repeated for all $N$ layers, progressively refining the motion representation at different temporal granularities. After $N$ iterations, we obtain the complete multi-scale token set $x=(x^{(1)}, x^{(2)}, \ldots, x^{(N)})$ for SAR modeling. The final dequantized embedding $\hat{z}$ fed into the detokenizer for motion reconstruction is computed as the sum of all intermediate upsampled embeddings: $\hat{z}=\sum_{n=1}^N \hat{z}^{(n)}$ (indicated by the addition symbol in Fig.~\ref{fig:whole_pipeline}). Listing~\ref{listing:code} presents the PyTorch-style pseudocode for this forward process.

\noindent\textbf{2) SAR Model Training Stage.}
Leveraging multi-scale tokens, SAR modeling reformulates traditional token-by-token prediction~\cite{zhang2023generating} into scale-by-scale generation. The blue dashed line in Fig.~\ref{fig:whole_pipeline} illustrates the SAR model training flow. Given input $(\texttt{[sos]}, x^{(1)}, x^{(2)}, \ldots, x^{(N-1)})$, SAR predicts $(x^{(1)}, x^{(2)}, \ldots, x^{(N)})$, where $x^{(n)} = (x^{(n)}_1, \dots, x^{(n)}_{s_n})$ represents the token sequence at scale $s_n$ (as defined in Eq.~\ref{eq:multi-scale}). The SAR likelihood is defined as:
\begin{equation}
p(x^{(1)}, \dots, x^{(N)} \mid t) = \prod_{n=1}^{N} p(x^{(n)} \mid x^{(1)}, \dots, x^{(n-1)}, t)
\end{equation}  
where all tokens in $x^{(n)}$ are predicted simultaneously at step $n$. A scale-wise causal attention mask ensures that each $x^{(n)}$ can only attend to $x^{(\leq n)}$, which makes the generation at each step condition on flattened features from previous scales and text embeddings $t$ (produced by the text encoder). Thanks to scale-wise autoregressive modeling, the inference steps align with the number of multi-scale quantizer layers $N$. Following MoSa, we set $N=10$ in our experiments. This reduces inference from 49 sequential steps (for token-by-token generation) to 10, since SAR predicts all tokens within each scale in parallel at each step. Moreover, the introduction of multi-scale intermediate representations enhances the model's contextual modeling capability.

\noindent\textbf{3) Generation Stage.} With both LG-Tok and the SAR model trained, they can be jointly leveraged to generate high-quality, diverse motions from natural language descriptions. The purple dashed line in Fig.~\ref{fig:whole_pipeline} illustrates the complete generation flow, comprising three key phases: autoregressive token generation, dequantization, and motion detokenization. The SAR model first predicts (samples) the first scale's token sequence  $\hat{x}^{(1)}$ in parallel, conditioned on text embedding $t$ extracted by the text encoder. Subsequently, the prediction of tokens $\hat{x}^{(2)}$ at the second scale is conditioned on $(\hat{x}^{(1)}, t)$; the prediction of tokens $\hat{x}^{(3)}$ at the third scale is conditioned on $(\hat{x}^{(1)}, \hat{x}^{(2)}, t)$, $\ldots$. This autoregressive process continues iteratively until all $N$ scales are predicted. Then, the predicted multi-scale tokens $(\hat{x}^{(1)},\ldots,\hat{x}^{(N)})$  are processed through dequantization. First, each scale's tokens are dequantized by getting their corresponding embeddings from the codebook. Following the upsampling and summation operations described in stage 1), these embeddings are combined to form the final dequantized embeddings $\hat{z}$. Finally, detokenization is performed: the learnable mask tokens in the LG-Tok detokenizer interact with both $\hat{z}$ and text embedding $t$ through separate cross-attention layers to decode the final motion output.

\section{Quantitative Evaluation on the KIT-ML Dataset}
\label{sec:kit_ml}

\input{table/table1_kit}

As shown in Table~\ref{tab:table1_kitml}, LG-Tok achieves state-of-the-art performance on the KIT-ML dataset, consistently outperforming competing methods across the majority of evaluation metrics.

\section{Feature Extractor Training on the Motion-X Dataset}
\label{sec:feature_extractor_training}

Quantitative evaluation of stochastic generative models has been well-established in the image generation domain. This typically involves extracting high-dimensional features from deep networks (\eg, Inception V3). These high-dimensional features are then used to measure distributional differences, diversity, and other properties. Similarly, quantitative evaluation for text-to-motion generation has been inspired by this paradigm. Guo \etal\cite{guo2022generating} trained separate feature extractors for text and motion using contrastive loss, and subsequent methods have adopted these extractors to evaluate key metrics such as FID and R-Precision. Recently, Meng \etal\cite{meng2025rethinking} revealed redundancy in data representations and retrained feature extractors on HumanML3D using tighter representations, while also introducing a MotionCLIP-style~\cite{tevet2022motionclip} network architecture for CLIP-based score evaluation. In our experiments, we adopt the HumanML3D feature extractor they provide and use it to reproduce representative methods. Since no feature extractor was provided for the larger Motion-X dataset, we trained a corresponding feature extractor following the discussions by Meng \etal in their GitHub issue\footnote{\url{https://github.com/neu-vi/MARDM/issues/8}}. Notably, we made two reasonable adaptations:

\noindent\textbf{1) Vocabulary handling.} We observe that the Motion-X dataset contains a significantly more diverse vocabulary. When using the original vocabulary provided by Guo \etal, many out-of-vocabulary words are uniformly marked as \texttt{unk} (unknown token), resulting in missing word embeddings. To better preserve and understand semantic information, we use GloVe 6B pre-trained word embeddings and perform part-of-speech tagging directly with spaCy.

\noindent\textbf{2) Batch size adjustment.}  When training the MotionCLIP-style network for CLIP-based score evaluation, we observe that mini-batch contrastive training (\eg, batch size = 8) results in negligible training loss reduction. This phenomenon likely stems from the inherently high similarity among motion descriptions, causing the negative sample set to contain numerous ``false negatives" that are semantically highly related to the anchor description, thereby undermining the learned embeddings~\cite{yan2023cross, petrovich2023tmr}. To mitigate this issue, we appropriately increase the batch size during training to enhance negative sample diversity. We ultimately set the batch size to 32 and conduct corresponding ablation studies:

\begin{table*}[!h]
\caption{Ablation study on batch size for training the MotionCLIP-style CLIP-score evaluator on Motion-X.}
\label{tab:batch}
\centering
\begin{tabular}{@{}c|c@{}}
\toprule
Batch size & CLIP-score \\ \midrule
8 & 0.012 \\
16 & 0.026 \\
32 & \textbf{0.672} \\
64 & 0.656 \\
128 & 0.630 \\ \bottomrule
\end{tabular}
\end{table*}

\medskip
For representation, we convert Motion-X to a body-only \texttt{meng} representation via the official scripts, dropping hand and facial keypoints. This choice is data-driven: only 2.74\% of Motion-X text samples contain hand- or face-related terms, so retaining those modalities would introduce domain mismatch with negligible semantic benefit while substantially complicating baseline adaptation.

\section{Details of Baseline Reproduction on the Motion-X Dataset}
\label{sec:baseline_reprod}

After training the Motion-X feature extractor, we reproduced representative baselines with extended training for optimal validation performance: StableMoFusion (150K steps vs. 50K original), MotionDiffuse (100 vs. 50 epochs), and MoMask tokenizer (100 vs. 50 epochs), \etc. Additionally, for tokenizer-based models (both discrete and continuous), we further enhance robustness by training on out-of-distribution (OOD) data. We observe that under the current generation paradigm targeting $\sim$10s of motions ($< 200$ frames), a substantial number of longer motion sequences ($\ge 200$ frames) remain unused. We leverage these as OOD data by scanning them with a sliding window of 200 frames for tokenizer training. Notably, in comparison to pre-training followed by fine-tuning, we employ parallel training: each batch comprises 50\% OOD samples and 50\% in-distribution samples. This prevents catastrophic forgetting while enabling knowledge transfer from extended motion sequences. Table~\ref{tab:ood} demonstrates the effectiveness of this approach, showing reconstruction improvements for LG-Tok and MoMask's tokenizer on the Motion-X test set:

\begin{table*}[!h]
\caption{Training on OOD data improves reconstruction performance for tokenizer-based models on the Motion-X test set.}
\label{tab:ood}
\centering
\renewcommand{\arraystretch}{1.2}
\resizebox{0.75\textwidth}{!}{%
\begin{tabular}{lcccccc}
\toprule
\multirow{2}{*}{Method} & \multicolumn{3}{c}{R Precision$\uparrow$} & \multirow{2}{*}{FID$\downarrow$} & \multirow{2}{*}{MultiModal Dist$\downarrow$} & \multirow{2}{*}{MPJPE$\downarrow$} \\ \cline{2-4}
 & Top 1 & Top 2 & Top 3 &  &  &  \\ \midrule
MoMask & \etb{0.554}{.002} & \etb{0.743}{.001} & \etb{0.829}{.001} & \etb{0.394}{.005} & \etb{4.284}{.006} & \textbf{24.9} \\
\,\, \textit{w/o} OOD training & \et{0.540}{.002} & \et{0.727}{.002} & \et{0.813}{.001} & \et{0.427}{.004} & \et{4.482}{.008} & 34.1 \\ \midrule
LG-Tok & \etb{0.577}{.002} & \etb{0.773}{.001} & \etb{0.858}{.001} & \etb{0.041}{.001} & \etb{3.890}{.008} & \textbf{31.0} \\
\,\, \textit{w/o} OOD training & \et{0.563}{.002} & \et{0.757}{.001} & \et{0.842}{.001} & \et{0.086}{.003} & \et{4.098}{.007} & 32.7 \\ \bottomrule
\end{tabular}
}
\end{table*}

\begin{figure}[!t]
\centering
\includegraphics[width=0.7\columnwidth]{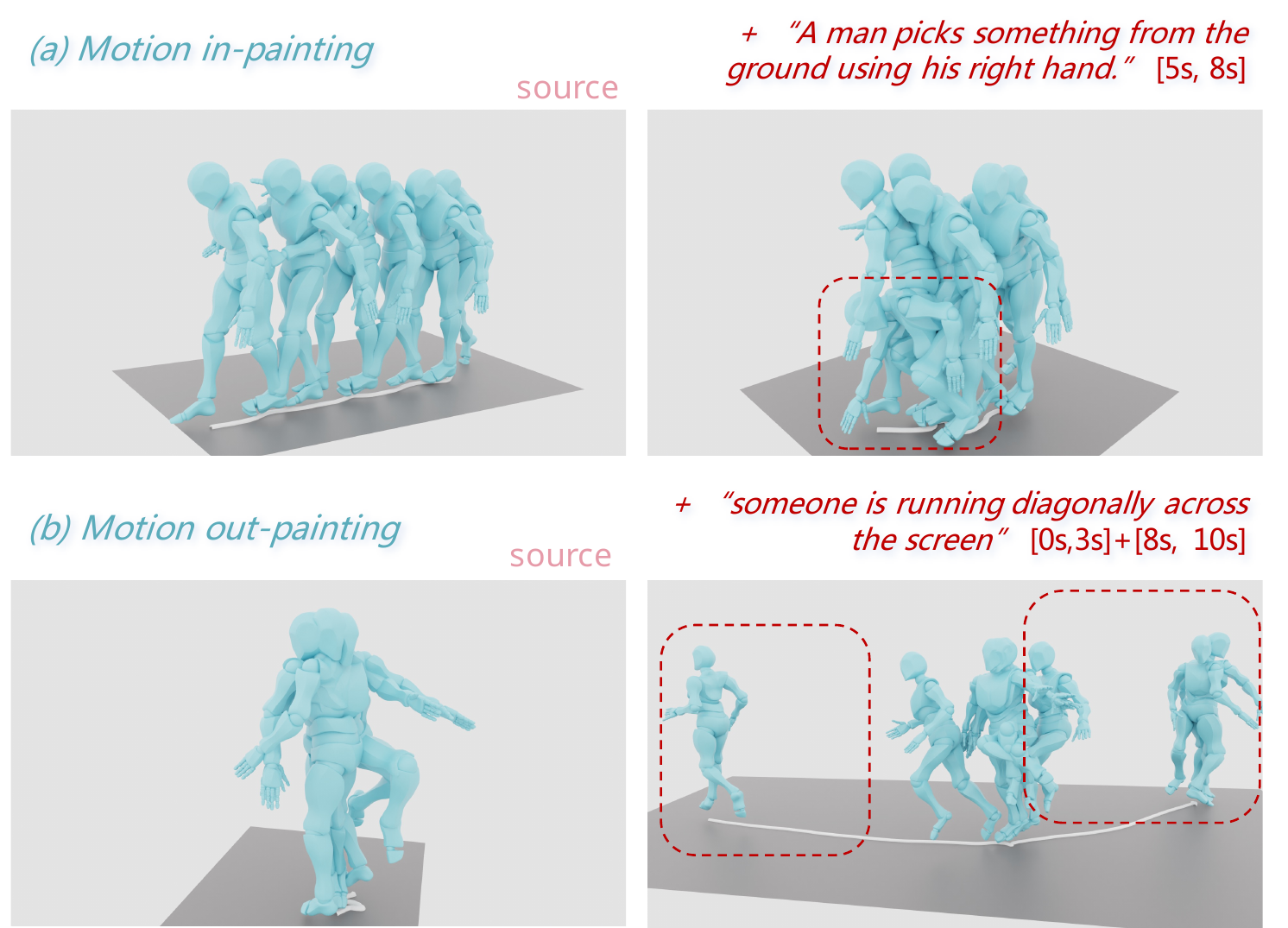}
\caption{\textbf{Motion editing results from LG-Tok.} (a) \textit{Motion in-painting}: given a source motion sequence with overlapping characters (left), our model generates a coherent edited motion where a man bends down and picks something from the ground using his right hand during $[5\text{s}, 8\text{s}]$ (red dashed box), seamlessly blending with the surrounding context. (b) \textit{Motion out-painting}: starting from a single source character (left), LG-Tok extends the motion temporally and spatially, creating smooth continuations at $[0\text{s}, 3\text{s}]$ (left red box) and synthesizing dynamic running motions diagonally across the scene at $[8\text{s}, 10\text{s}]$ (right red box).}
\label{fig:motion_editing}
\end{figure}

\section{Application: Motion Editing}
\label{sec:motion_editing}

We provide motion editing results to demonstrate that LG-Tok's benefits extend beyond standard text-to-motion generation to constrained tasks. Following protocols in prior generative models (\eg, MoSa~\cite{liu2025mosa}, MoMask~\cite{guo2024momask}), while preserving the inherent editability of the generative model, we can further enhance precise control by applying an edit mask during the \textit{detokenization} stage to enable language guidance only in designated temporal regions. Specifically, given a binary temporal mask $\mathcal{M} \in \{0,1\}^{F}$ defining the regions to edit, the detokenizer applies text guidance conditionally:
\begin{equation}
\hat{m} = \mathcal{D}(\hat{m}_{l}, \hat{z}, t, \mathcal{M})
\end{equation}
where mask tokens interact with text embeddings $t$ through cross-attention only in regions where $\mathcal{M}=1$, while regions with $\mathcal{M}=0$ rely solely on dequantized embeddings $\hat{z}$ without text conditioning. This enables precise temporal control for partial generation tasks such as motion editing and in-betweening. As illustrated in Fig.~\ref{fig:motion_editing}, our model successfully performs both motion in-painting and out-painting, seamlessly blending edited regions with the surrounding motion context. These results demonstrate LG-Tok's capability to maintain motion coherence while adapting to diverse text-guided editing scenarios.

\section{Evaluation of Average Inference Time}
\label{sec:average_inference_time}

\begin{table}[!h]
\caption{Average Inference Time Results Comparison between our method and baseline methods.}
\label{tab:AIT}
\centering
\resizebox{1.0\textwidth}{!}{
\begin{tabular}{cccccccc}
\toprule
MotionDiffuse~\cite{zhang2022motiondiffuse} & T2M-GPT~\cite{zhang2023generating} & MoMask~\cite{guo2024momask} & MMM~\cite{pinyoanuntapong2024mmm} & StableMoFusion~\cite{huang2024stablemofusion} & MoSa~\cite{liu2025mosa} & MARDM~\cite{meng2025rethinking} & LG-Tok (Ours) \\ \midrule
4.086s & 0.127s & 0.062s & 0.085s & 0.153s & 0.045s & 2.109s & 0.121s \\ \bottomrule
\end{tabular}}
\end{table}

To provide a comprehensive evaluation, we assess the computational efficiency of our method in terms of average inference time (AIT). Table~\ref{tab:AIT} reports the efficiency of motion generation over 100 samples on a single Nvidia 4090 device. While LG-Tok achieves superior generation quality across all metrics, it maintains millisecond-scale latency (0.121s), demonstrating an effective balance between generation quality and computational efficiency.

\section{Qualitative Comparison for Text Guidance}
\label{sec:qualitative_comparsion_for_text_guidance}

\begin{figure*}[!h]
    \centering
    \includegraphics[width=1.0\textwidth, trim=40mm 0mm 30mm 0mm, clip]{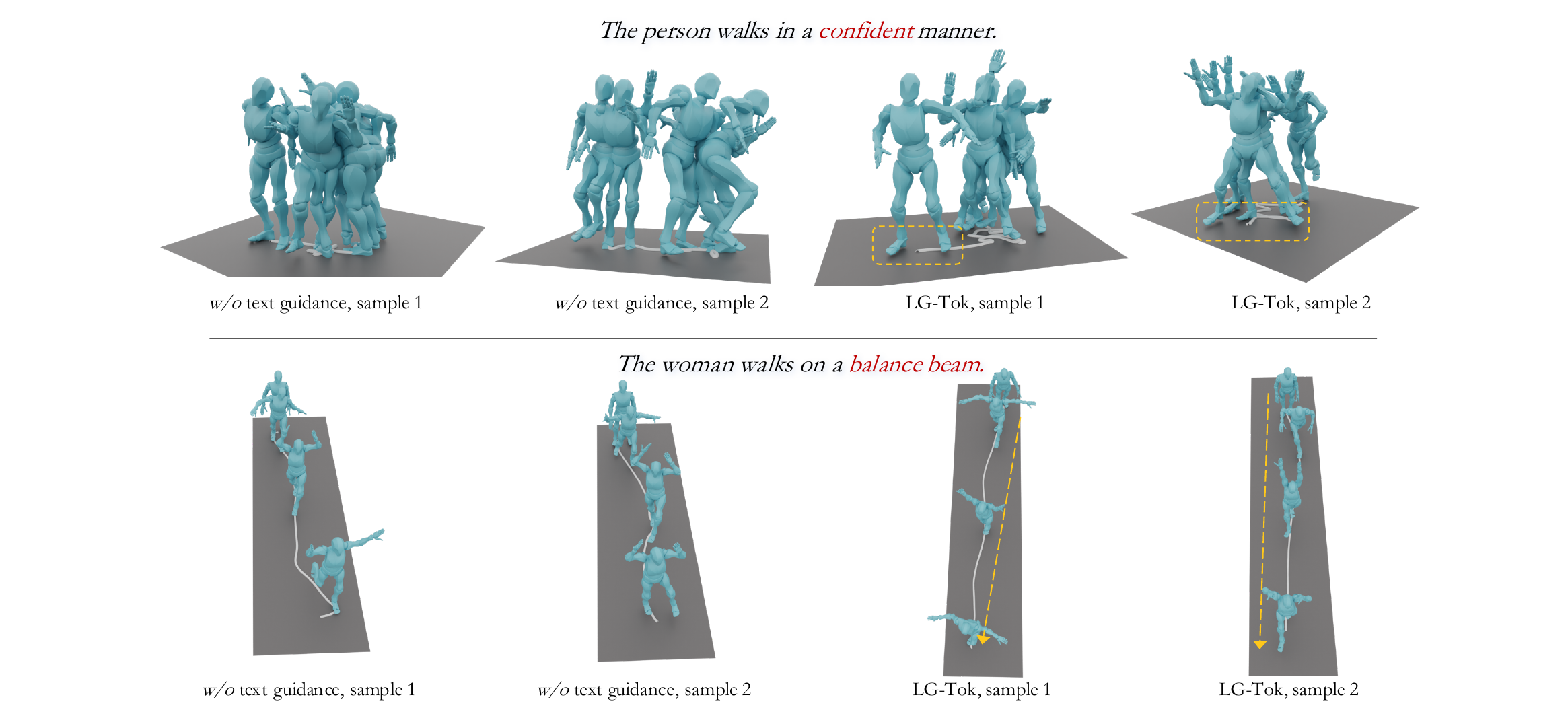}
    \caption{\textbf{Qualitative comparison for text guidance.} Qualitative comparison based on Motion-X checkpoint demonstrating that text-guided tokenization (LG-Tok) generates motions with improved semantic alignment to text descriptions compared to training without (w/o) text guidance.}
    \label{fig:text_comparasion}
\end{figure*}

\noindent We compare the qualitative results of our transformer-based tokenizer with text guidance (LG-Tok) against the variant without text guidance (w/o text guidance) on the Motion-X dataset. Fig.~\ref{fig:text_comparasion} illustrates two representative examples. We argue that incorporating language guidance enables LG-Tok to capture high-level semantic representations, which better align the generated motions with textual conditions. In the first example (``\textit{The person walks in a confident manner}"), both variants generate walking motions. Yet, the \textit{w/o} text guidance version exhibits predominantly single-foot movements with noticeable foot-slip artifacts. In contrast, LG-Tok produces bilateral swaying motions that better convey the semantic notion of ``confident" walking. In the second example (``\textit{The woman walks on a balance beam}"), LG-Tok generates a nearly straight trajectory, while \textit{w/o} text guidance version shows slight curvature deviations. These qualitative comparisons demonstrate the advantages of integrating natural language during tokenization. The role of language guidance in the detokenization stage is further analyzed in the next section.

\begin{figure*}[!ht]
    \centering
    \includegraphics[width=1.0\textwidth, trim=28mm 0mm 40mm 0mm, clip]{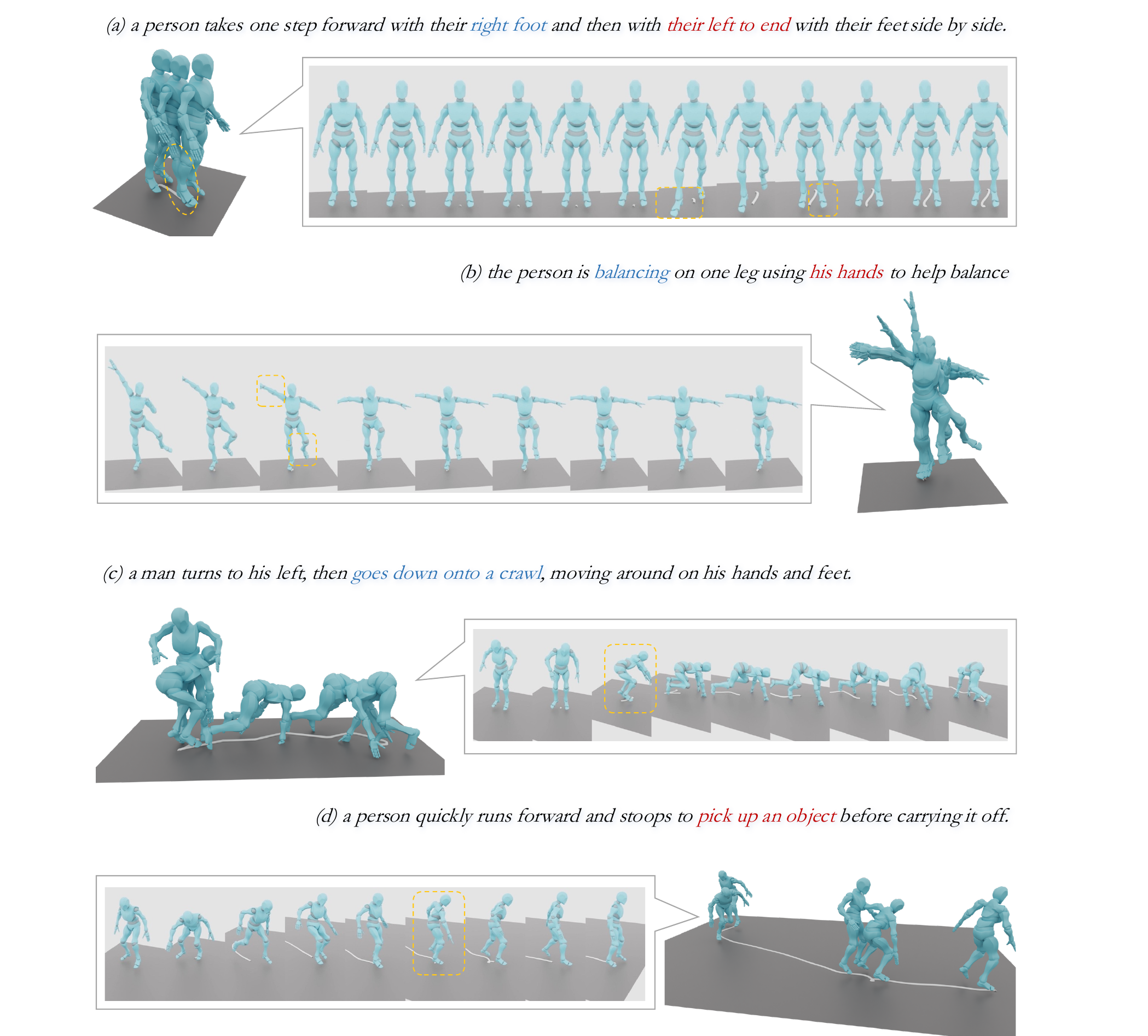}
    \caption{\textbf{More qualitative results on HumanML3D.} Each motion sequence is visualized with temporal expansion, with yellow dashed boxes highlighting key frames corresponding to the emphasized actions in the text descriptions (shown in blue and red).}
    \label{fig:more_qualitative} 
\end{figure*}

\section{The Role of Language Guidance in the Detokenization Stage during Generation}
\label{sec:role_of_language}

\begin{figure}[!h]
    \centering
    \includegraphics[width=0.34\linewidth]{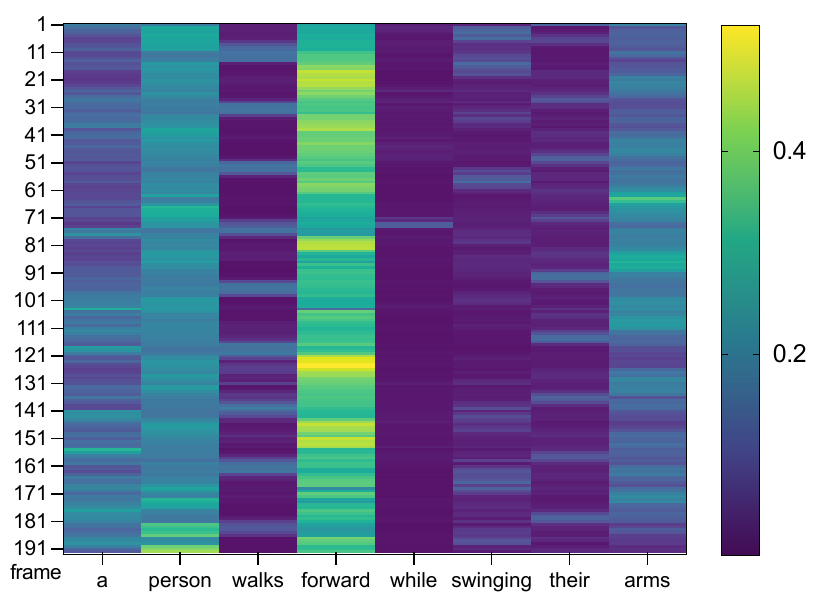}\quad\quad\quad\quad\quad\quad
    \includegraphics[width=0.34\linewidth]{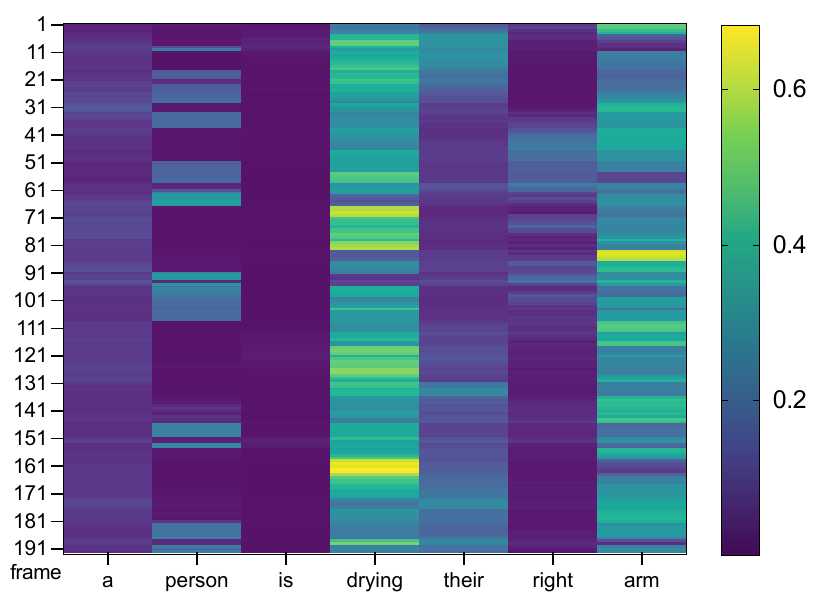}
    \caption{\textbf{Cross-attention weight visualization in the detokenizer.} Each motion frame attends most strongly to semantically pivotal words (\eg, \textit{forward}, \textit{walks}; \textit{drying}, \textit{arm}), confirming word-level text-to-motion alignment.}
    \label{fig:heatmap}
\end{figure}

Fig.~\ref{fig:heatmap} directly visualizes cross-attention weights in the detokenizer: each row corresponds to a motion frame and each column to a text token, with brighter cells indicating stronger attention. In both examples, motion frames attend most strongly to semantically pivotal words---frames during a forward stride peak on \textit{forward} and \textit{walks}, while arm-drying frames peak on \textit{drying} and \textit{arm}---confirming that language-guided tokenization induces precise word-level text-to-motion alignment.

Our language-guided tokenization establishes motion-text alignment during training, enabling the detokenizer to act as a \textit{semantic amplifier} during generation. As shown in Table~\ref{tab:language_decoding}, language-free guidance decoding consistently improves generation quality across both datasets. This phenomenon reveals that language guidance in detokenization serves as a semantic refinement mechanism: the tokens generated by the generative model already encode high-level semantic information learned during tokenization, and the detokenizer's language conditioning further enhances this semantic coherence. Notably, even in free-form generation (where the generative model operates without text conditioning), language-free guidance decoding also improves quality (FID: 4.025$\to$3.564 on HumanML3D, 25.093$\to$14.715 on Motion-X), demonstrating that the detokenizer can leverage its learned language-motion alignment to refine arbitrary token sequences semantically. This validates our hypothesis that language guidance in detokenization acts as a continuous semantic enhancement process, translating compact token representations into motion sequences with improved semantic fidelity.

\begin{table*}[!h]
\caption{Effect of language-free guidance decoding during generation.} \label{tab:language_decoding}
\centering
\renewcommand{\arraystretch}{1.2}
\resizebox{0.85\textwidth}{!}{
\begin{tabular}{l|cc|cc|cc}
\toprule
\multirow{2}{*}{Method} & \multicolumn{2}{c|}{FID $\downarrow$} & \multicolumn{2}{c|}{R-Precision Top-1 $\uparrow$} & \multicolumn{2}{c}{CLIP-score $\uparrow$} \\
& HumanML3D & Motion-X & HumanML3D & Motion-X & HumanML3D & Motion-X \\
\midrule
\multicolumn{7}{l}{Free-form Generation (no text in generative model)} \\
\quad \textit{w/o} language decoding & \et{4.025}{.094} & \et{25.093}{.322} & \et{0.044}{.001} & \et{0.048}{.001} & \et{0.089}{.002} & \et{0.070}{.001} \\
\quad \textit{w/} language decoding & \etb{3.564}{.079} & \etb{14.715}{.200} & \etb{0.054}{.001} & \etb{0.101}{.001} & \etb{0.108}{.002} & \etb{0.154}{.001} \\
\midrule
\multicolumn{7}{l}{Text-conditioned Generation (standard setting)} \\
\quad \textit{w/o} language decoding & \et{0.061}{.003} & \et{0.139}{.008} & \et{0.534}{.003} & \et{0.576}{.002} & \et{0.665}{.001} & \et{0.680}{.000} \\
\quad \textit{w/} language decoding & \etb{0.057}{.003} & \etb{0.088}{.006} & \etb{0.542}{.003} & \etb{0.582}{.002} & \etb{0.669}{.001} & \etb{0.682}{.000} \\
\bottomrule
\end{tabular}
}
\end{table*}

\section{More Qualitative Results}
\label{sec:more_qualitative_comparison}

We generate additional qualitative results using our model trained on HumanML3D to demonstrate the effectiveness of our approach further. As shown in Fig.~\ref{fig:more_qualitative}, each human motion is accompanied by a corresponding temporal expansion visualization, with key frames highlighted in yellow dashed boxes to emphasize the critical motion phases described in the text prompts. In example (a), the model accurately captures the sequential stepping motion, with the \textit{right foot} movement clearly visible in the early frames (highlighted) and the subsequent \textit{left foot} action leading to the final pose with \textit{feet side by side}. Example (b) demonstrates precise control over complex balance motions, where the model generates realistic \textit{balancing} poses on one leg while correctly positioning the \textit{hands} (highlighted) to maintain equilibrium. In example (c), the generated motion faithfully follows the described sequence: the person first \textit{turns to his left}, then smoothly transitions into a crawling position (\textit{goes down onto a crawl}, highlighted), and continues moving on hands and feet. Example (d) showcases the model's ability to synthesize fine-grained actions, where the person \textit{quickly runs forward}, executes a precise stopping motion, and performs the detailed action of \textit{picking up an object} (highlighted) before carrying it away. These results demonstrate that our language-guided tokenization enables the model to generate motions that are not only semantically aligned with text descriptions but also exhibit smooth temporal transitions and accurate correspondence between linguistic emphasis and motion details.

\section{Limitations}
\label{sec:limitations}

Unlike convolutional tokenizers (\eg, MoMask with 64-frames window size requiring $\sim$2GB memory), LG-Tok's attention-based architecture with global attention over motion, text, and latent tokens consumes $10\times$ more memory. To mitigate this limitation, we introduce mixed-precision training, which enables tokenizer training on a single 24GB GPU (RTX 4090). However, this increased memory footprint inevitably extends training time: tokenizer training requires approximately 2 days on a single RTX 4090, compared to several hours for convolutional baselines. For inference efficiency analysis, please refer to Sec.~\ref{sec:average_inference_time}. Despite these computational costs, the substantial quality improvements justify this overhead.

\begin{listing}[htbp]
\begin{lstlisting}[language=Python, 
    basicstyle=\ttfamily\scriptsize,
    keywordstyle=\color{blue}\bfseries,
    commentstyle=\color{red!70!black},
    stringstyle=\color{red!70!black},
    numbers=left,
    numberstyle=\tiny\color{black},
    stepnumber=1,
    numbersep=10pt,
    frame=none,
    breaklines=true,
    breakatwhitespace=false,
    showstringspaces=false,
    tabsize=4,
    xleftmargin=1em,
    captionpos=b]
import torch.nn as nn
from torch.nn import functional as F

class LGTok(nn.Module):
    def __init__(self, scales=(2, 4, 6, 9, 12, 16, 20, 25, 30, 36), **kwargs):
        super(LGTok, self).__init__()
        # The predefined scale scheduler S = (s_1, s_2, ..., s_N)
        self.scales = scales 
        self.N = len(scales)
        self.tokenizer =  Tokenizer(**kwargs)
        self.detokenizer =  Detokenizer(**kwargs)
        self.quantizers = nn.ModuleList([Quantizer(**kwargs) for _ in range(self.N)])
        
    def forward(self, m, t):
        # Forward pass for the LG-Tok model.
        # Args: m: input motion, t: natural language
        # Returns: m_hat: reconstructed motion, x: multi-scale token set
        z = self.tokenizer(m, t)
        z_hat, s_N = 0., self.scales[-1]
        # Eq (5) in paper, multi-scale token set x = (x^(1), x^(2), ..., x^(N))
        x = []
        
        # Multi-scale quantization
        for n in range(self.N):
            s_n, quantizer = self.scales[n], self.quantizers[n]

            # Perform downsampling and quantization: x^(n) = Q^(n)(downsample(z^(n), s_n))
            x_n = quantizer(F.interpolate(z, size=(s_n))) 

            # Get corresponding codebook entries, form the dequantized embeddings
            z_prime_hat_n = quantizer.codebook(x_n)

            # Perform upsampling to bring the embedding scale back to s_N
            z_hat_n = F.interpolate(z_prime_hat_n, size=(s_N)) 

            # Compute the residual for the next quantization layer
            z = z - z_hat_n 
            z_hat += z_hat_n # z_hat = sum of all z_hat^(n)
            x.append(x_n)

        m_hat = self.detokenizer(z_hat, t)
        return m_hat, x
\end{lstlisting}
\caption{\textbf{PyTorch-style pseudocode for multi-scale quantization in LG-Tok.} The tokenizer encodes motion with text guidance into latent features $z$. Multi-scale quantization iteratively performs downsample-quantize-upsample operations across $N$ scales, producing multi-scale token set $x=(x^{(1)}, \ldots, x^{(N)})$ for SAR training. The final dequantized embeddings $\hat{z}=\sum_{n=1}^N \hat{z}^{(n)}$ are decoded by the detokenizer to reconstruct motion $\hat{m}$.}
\label{listing:code}
\end{listing}

%% file: table/table1_kit.tex
\begin{table*}[htbp]
\centering
\caption{\textbf{Quantitative evaluation on the 
KIT-ML test set.} Each experiment is
evaluated 20 times, and $\pm$ indicates a 95\% confidence interval. \textbf{Bold} and \underline{underline} indicate the best and the second best result, respectively. `$\dagger$' denotes our reimplementation on the \texttt{meng} representation~\cite{meng2025rethinking}.}
\label{tab:table1_kitml}
\renewcommand{\arraystretch}{1.3}
\resizebox{1.0\textwidth}{!}{%
\begin{tabular}{@{}llcccccccc@{}}
\toprule
\multirow{2}{*}{Methods} & \multirow{2}{*}{Venues} & \multirow{2}{*}{\#Tokens} & \multicolumn{3}{c}{R Precision$\uparrow$} & \multirow{2}{*}{FID$\downarrow$} & \multirow{2}{*}{MultiModal Dist$\downarrow$} & \multirow{2}{*}{MultiModality$\uparrow$} & \multirow{2}{*}{CLIP-score$\uparrow$} \\ \cmidrule(lr){4-6}
&  &  & Top 1 & Top 2 & Top 3 &  &  &  &  \\ \midrule
\textbf{Real motions} & \quad- & - & \et{0.374}{.006} & \et{0.605}{.006} & \et{0.746}{.006} & \et{0.000}{.000} & \et{3.322}{.013} & - & \et{0.696}{.001} \\ 
\textbf{Our LG-Tok (Recons.)} & \quad- & - & \et{0.373}{.005} & \et{0.595}{.006} & \et{0.734}{.004} & \et{0.108}{.003} & \et{3.395}{.011} & - & - \\ \midrule
 MotionDiffuse~\cite{zhang2022motiondiffuse} & TPAMI\textquotesingle\,24 & - & \et{0.344}{.009} & \et{0.536}{.007} & \et{0.658}{.007} & \et{3.845}{.087} & \et{4.167}{.054} & \et{1.774}{.217} & \et{0.626}{.006} \\
 MLD~\cite{chen2023executing} & CVPR\textquotesingle\,23 & - & \et{0.351}{.007} & \et{0.536}{.007} & \et{0.658}{.007} & \et{0.492}{.047} & \et{3.746}{.044} & \ets{1.803}{.164} & \et{0.646}{.006} \\
  T2M-GPT~\cite{zhang2023generating} & CVPR\textquotesingle\,23 & 49 & \et{0.359}{.007} & \et{0.553}{.007} & \et{0.690}{.013} & \et{0.593}{.053} & \et{3.765}{.046} & \et{1.798}{.157} & \et{0.651}{.005} \\
 ReMoDiffuse~\cite{zhang2023remodiffuse} & ICCV\textquotesingle\,23 & - & \et{0.356}{.004} & \et{0.572}{.007} & \et{0.706}{.009} & \et{1.725}{.053} & \et{3.735}{.036} & \etb{1.928}{.127} & \et{0.665}{.005} \\
 MMM~\cite{pinyoanuntapong2024mmm} & CVPR\textquotesingle\,24 & 49 & \et{0.363}{.005} & \et{0.569}{.006} & \et{0.724}{.006} & \et{0.478}{.034} & \et{3.629}{.028} & \et{1.455}{.106} & \et{0.660}{.003} \\
 MoMask~\cite{guo2024momask} & CVPR\textquotesingle\,24 & 294 & \et{0.369}{.005} & \et{0.588}{.005} & \et{0.731}{.005} & \et{0.411}{.026} & \et{3.577}{.021} & \et{1.309}{.058} & \et{0.669}{.002} \\
 MARDM~\cite{meng2025rethinking} & CVPR\textquotesingle\,25 & - & \ets{0.387}{.006} & \ets{0.610}{.006} & \ets{0.749}{.006} & \ets{0.242}{.014} & \ets{3.374}{.019} & \et{1.312}{.053} & \ets{0.692}{.002} \\
 \textbf{Our LG-Tok (Gen.)} & \quad- & 236 & \etb{0.401}{.006} & \etb{0.626}{.005} & \etb{0.765}{.006} & \etb{0.185}{.008} & \etb{3.270}{.013} & \et{1.257}{.067} & \etb{0.702}{.002} \\ \bottomrule
\end{tabular}%
}
\end{table*}